%% file: pan2025rss_arxiv_v2.tex
\documentclass[conference]{IEEEtran} 
\usepackage{times}
\input{stachnisslab-latex}

\input{stachnisslab-math}

\usepackage{multicol}
\usepackage{subfiles}
\usepackage{booktabs}
\usepackage{siunitx}
\usepackage{multirow}
\usepackage{pifont}
\usepackage{amssymb}
\usepackage{amsmath}
\usepackage{subfigure}
\usepackage{tabularx}
\usepackage{comment}
\usepackage{mathtools} 
\usepackage{makecell}
\usepackage{tabularx} 
\usepackage{color}
\usepackage{colortbl}
\usepackage[dvipsnames]{xcolor}
\usepackage[numbers, sort, compress]{natbib}
\usepackage[bookmarks=true]{hyperref}  

\renewcommand{\and}{\hspace{1.0cm}} 

\begin{document}

\title{PINGS: Gaussian Splatting Meets Distance Fields within a Point-Based Implicit Neural Map}

\author{Yue Pan$^*$ \qquad Xingguang Zhong$^*$ \qquad 
Liren Jin$^*$ \qquad Louis Wiesmann$^*$\\
Marija Popovi\'{c}$^{\ddagger}$ \qquad Jens Behley$^*$ \qquad Cyrill Stachniss$^{*, \dagger}$\\[2mm]
$^*$ Center for Robotics, University of Bonn, Germany\\
$^\ddagger$ MAVLab, TU Delft, the Netherlands\\
$^{\dagger}$ Lamarr Institute for Machine Learning and Artificial Intelligence, Germany
}

\twocolumn[{%
\renewcommand\twocolumn[1][]{#1}%

\maketitle



\begin{center}
  \centering
  \vspace{-14pt} 
  \captionsetup{type=figure}
  \includegraphics[width=0.99\linewidth]{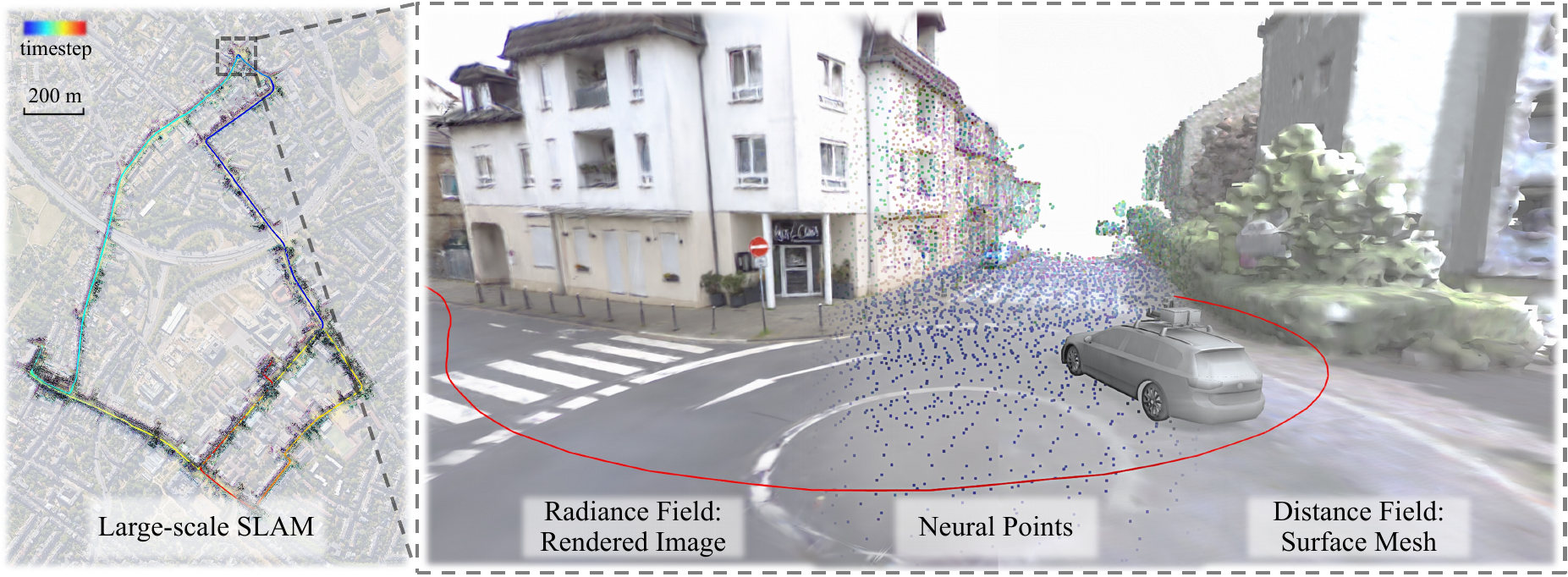}
  \setlength{\abovecaptionskip}{5pt}
  \captionof{figure}{
  We present PINGS, a novel LiDAR-visual SLAM system unifying distance field and radiance field mapping using an elastic point-based implicit neural representation. 
  On the left, we show a globally consistent neural point map overlaid on a satellite image. 
  The map was built using PINGS from around 10,000 LiDAR scans and 40,000 images collected by a robot car driving in an urban environment for around \SI{5}{\kilo\meter}. 
  %
  The estimated trajectory is overlaid on the map and colorized according to the timestep.
  On the right, we show a zoomed-in view of a roundabout mapped by PINGS. 
  It illustrates from left to right the rendered image from the Gaussian splatting radiance field, neural points colorized by the principal components of their geometric features, and the reconstructed mesh from the distance field (colorized by the radiance field). 
  The red line indicates the local trajectory of the robot car (shown as the CAD model).
  }
  \label{fig:teaser}
  \vspace{0pt}
\end{center}%
}]


\begin{abstract}
  Robots benefit from high-fidelity reconstructions of their environment, which should be geometrically accurate and photorealistic to support downstream tasks.
  While this can be achieved by building distance fields from range sensors and radiance fields from cameras, realising scalable incremental mapping of both fields consistently and at the same time with high quality is challenging. In this paper, we propose a novel map representation that unifies a continuous signed distance field and a Gaussian splatting radiance field within an elastic and compact point-based implicit neural map. By enforcing geometric consistency between these fields, we achieve mutual improvements by exploiting both modalities. We present a novel LiDAR-visual SLAM system called PINGS using the proposed map representation and evaluate it on several challenging large-scale datasets. Experimental results demonstrate that PINGS can incrementally build globally consistent distance and radiance fields encoded with a compact set of neural points. Compared to state-of-the-art methods, PINGS achieves superior photometric and geometric rendering at novel views by constraining the radiance field with the distance field. Furthermore, by utilizing dense photometric cues and multi-view consistency from the radiance field, PINGS produces more accurate distance fields, leading to improved odometry estimation and mesh reconstruction.
  We also provide an open-source implementation of PINGS.
  %
\end{abstract}

\section{Introduction}
\label{sec:intro}


The ability to perceive and understand the surroundings is fundamental for autonomous robots.
At the core of this capability lies the ability to build a map --- a digital twin of the robot's workspace that is ideally both geometrically accurate and photorealistic, enabling effective spatial awareness and operation of the robot~\cite{mascaro2024annurevcontrol, hughes2024ijrr}.

Previous works in robotics mainly focus on the incremental mapping of an occupancy grid or a distance field using range sensors, such as LiDAR or depth cameras, which enable localization~\cite{grisetti2007tro}, collision avoidance~\cite{fox1997jra}, or exploration~\cite{stachniss2005rss}.
Recently, PIN-SLAM~\cite{pan2024tro} demonstrated that a compact point-based implicit neural representation can effectively model a continuous signed distance field (SDF) for LiDAR simultaneous localization and mapping~(SLAM), enabling both accurate localization and globally consistent mapping.

However, occupancy voxel grids~\cite{grisetti2007tro}, occupancy fields~\cite{zhao2022rss-occslam}, or distance fields~\cite{ortiz2022rss, pan2024tro} fall short of providing photorealistic novel view rendering of the scene, which is crucial for applications requiring dense photometric information.
This capability can be achieved by building an additional radiance field with visual data using representations such as neural radiance field (NeRF)~\cite{mildenhall2020eccv} or a \mbox{3D Gaussian splatting (3DGS)} model~\cite{kerbl2023tog-3dgs}.
%
%
Recent works demonstrated the potential of radiance fields, especially 3DGS, for various robotic applications including human-robot interaction~\cite{wildersmith2024iros}, scene understanding~\cite{zuo2024fmgs, zheng2024ral-gaussiangrasper}, simulation or world models for robotics learning~\cite{driess2022neurips-nerfrl, yang2023cvpr-unisim, abou-chakra2024corl}, visual localization~\cite{matsuki2024cvpr-monogs, bortolon2024eccv-6dgs}, and active reconstruction~\cite{jin2024iros-gsplanner, jin2025ral-activegs}.
Nevertheless, these approaches often assume well-captured image collections in bounded scenes with offline processing, limiting their applicability for mobile robotic applications.
Besides, radiance fields are not necessarily geometrically accurate, which can lead to issues in localization or planning.




%
In this paper, we investigate how to simultaneously build consistent, geometrically accurate, and photorealistic radiance fields as well as accurate distance fields for large-scale environments using LiDAR and camera data.
Building upon PIN-SLAM's~\cite{pan2024tro} point-based neural map for distance fields and inspired by Scaffold-GS~\cite{lu2024cvpr-scaffoldgs}, we propose a novel point-based model that additionally represents a Gaussian splatting radiance field.
By enforcing mutual supervision between these fields during incremental mapping, we achieve both improved rendering quality from the radiance field and more accurate distance field for better localization and surface reconstruction.
%


%

The main contribution of this paper is a novel LiDAR-visual SLAM system, called PINGS, that incrementally builds continuous SDF and Gaussian splatting radiance fields by exploiting their mutual consistency within a point-based neural map.
The distance field and radiance field infered from the elastic neural points enable robust pose estimation while maintaining global consistency through loop closure correction.
The compact neural point map can be efficiently stored and loaded from disk, allowing accurate surface mesh reconstruction from the distance field and high-fidelity real-time novel view rendering from the radiance field, as shown in~\figref{fig:teaser}.


In sum, we make four key claims:
%
(i) PINGS achieves better RGB and geometric rendering at novel views by constraining the Gaussian splatting radiance field using the signed distance field;
(ii) PINGS builds a more accurate signed distance field for more accurate localization and surface reconstruction by leveraging dense photometric cues from the radiance field; 
(iii) PINGS enables large-scale globally consistent mapping with loop closures; 
%
(iv) PINGS builds a more compact map than previous methods for both radiance and distance fields.
%


Our open-source implementation of PINGS is publicly available at: \mbox{\url{https://github.com/PRBonn/PINGS}}.

\section{Related Work}
\label{sec:related}


\subsection{Point-based Implicit Neural Representation}

Robotics has long relied on explicit map representations with discrete primitives like point clouds~\cite{zhang2014rss}, surfels~\cite{whelan2015rss, behley2018rss}, meshes~\cite{vizzo2021icra}, or voxel grids~\cite{hornung2013ar,newcombe2011ismar} for core tasks like localization~\cite{thrun2001ai} and planning~\cite{stachniss2005rss}.

Recently, implicit neural representations have been proposed to model radiance fields~\cite{mildenhall2020eccv} and geometric (occupancy or distance) fields~\cite{mescheder2019cvpr, park2019cvpr, ortiz2022rss} using \mbox{multi-layer perceptrons (MLP)}. 
These continuous representations offer advantages like compact storage, and better handling of regions with sparse observations or occlusions, while supporting conversion to explicit representations for downstream tasks.

Instead of using a single MLP for the entire scene, recent methods use hybrid representations that combine local feature vectors with a shared shallow MLP.
Point-based implicit neural representations~\cite{xu2022cvpr-pointnerf, pan2024tro} store optimizable features in a neural point cloud, which has advantages over grid-based alternatives through its flexible spatial layout and inherent elasticity under transformations for example caused by loop closures.
Point-based implicit neural representations have been used for modeling either radiance fields or distance fields for various applications including differentiable rendering~\cite{xu2022cvpr-pointnerf, chen2023iccv-neurbf}, dynamic scene modeling~\cite{abou-chakra2024wacv}, surface reconstruction~\cite{li2022cvpr-dccdif}, visual odometry~\cite{sandstrom2023iccv-pointslam,zhang2024eccv-glorie}, and globally consistent mapping~\cite{pan2024tro}. 
For example, PIN-SLAM~\cite{pan2024tro} effectively represents local distance fields with neural points for odometry estimation and uses the elasticity of these neural points during loop closure correction.

In this paper, we propose a novel LiDAR-visual SLAM system that is built on top of PIN-SLAM~\cite{pan2024tro} and encodes a Gaussian splatting radiance field within neural points while jointly optimizing it alongside the distance field.
Compared to NeRF-based approaches~\cite{xu2022cvpr-pointnerf, chen2023iccv-neurbf}, this offers faster novel view rendering suitable for robotics applications.

\subsection{Gaussian Splatting Radiance Field}

NeRF~\cite{mildenhall2020eccv} pioneered the use of MLPs to map 3D positions and view directions to color and volume density, encoding radiance fields through volume rendering-based training with posed RGB images.
More recently, 3DGS~\cite{kerbl2023tog-3dgs} introduced explicit 3D Gaussian primitives to represent the radiance fields, achieving high-quality novel view synthesis.
Compared to NeRF-based methods, 3DGS is more efficient by using primitive-based differentiable rasterization~\cite{wang2019tog-dss} instead of ray-wise volume rendering.
The explicit primitives also enables editing and manipulation of the radiance field.
These properties make 3DGS promising for robotics applications~\cite{jin2025ral-activegs, li2024arxiv-activesplat, matsuki2024cvpr-monogs, wildersmith2024iros, abou-chakra2024corl}.
However, two main challenges limit its usage: geometric accuracy and scalability for incremental mapping.
We discuss the related works addressing geometric accuracy in the following and addressing scalable mapping in \secref{subsec:large_scale_3d_reconstruction}.

While 3DGS achieves high-fidelity photorealistic rendering, it often lacks the geometric accuracy. 
%
To tackle this limitation, SuGaR~\cite{guedon2024cvpr-sugar} uses a hybrid representation to extract meshes from 3DGS and align the Gaussian primitives with the surface meshes.
To address the ambiguity in surface description, another solution is to flatten the 3D Gaussian ellipsoids to 2D disks~\cite{huang2024siggraph-2dgs, dai2024siggraph-gaussian-surfels, zhang2024arxiv-radegs, jiang2024arxiv-ligs}.
The 2D disks gradually align with surfaces during training, enabling more accurate depth and normal rendering.
However, extracting surface meshes from these discrete primitives still requires either TSDF fusion~\cite{newcombe2011ismar} with rendered depth or Poisson surface reconstruction~\cite{kazhdan2013acmgraphics}. 
%

Another line of works~\cite{yu2024tog-gof, song2024neurips-gvkf} model discrete Gaussian opacity as a continuous field, similar to NeRF-based surface reconstruction~\cite{wang2021neurips}.
Several works~\cite{chen2023arxiv-neusg, lyu2024tog-3dgsr, yu2024neurips-gsdf} jointly train a distance field with 3DGS and align the Gaussian primitives with the zero-level set of the distance field to achieve accurate surface reconstruction.
However, these methods rely solely on image rendering supervision for both 3DGS and neural SDF training without direct 3D geometric constraints, leading to ambiguities in textureless or specular regions.
The volume rendering-based SDF training also impacts efficiency.

While 3DGS originally uses structure-from-motion point clouds, robotic platforms with LiDAR can initialize primitives directly from LiDAR measurements~\cite{cui2024tog-letsgo, hong2024ral-livgaussmap, xie2024arxiv-gslivm}.
Direct depth measurements can further supervise depth rendering to improve geometric accuracy and convergence speed~\cite{matsuki2024cvpr-monogs, jiang2024arxiv-ligs}.

Our approach uniquely combines geometrically consistent 2D Gaussian disks with a neural distance field supervised by direct LiDAR measurements, enforcing mutual geometric consistency between the representations.
This differs from GSFusion~\cite{wei2024ral-gsfusion}, which maintains decoupled distance and radiance fields without mutual supervision.

\subsection{Large-Scale 3D Reconstruction}
\label{subsec:large_scale_3d_reconstruction}

This paper focuses on online large-scale 3D reconstruction.
%
There have been numerous works for the scalable occupancy or distance field mapping in the past decade, using efficient data structures such as an Octree~\cite{hornung2013ar, zhong2023icra}, voxel hashing~\cite{klingensmith2015rss, oleynikova2017iros, zhong2024cvpr}, an VDB~\cite{vizzo2022sensors, wu2024ral-vdbgpdf}, or wavelets~\cite{reijgwart2023rss-wavemap}.

Scalable radiance field mapping has also made significant progress recently.
For large scale scenes captured by aerial images, recent works~\cite{lu2024cvpr-scaffoldgs, ren2024arxiv-octreegs, liu2024eccv-citygaussian} demonstrate promising results using level-of-detail rendering and neural Gaussian compression.
For driving scenes with short sequences containing hundreds of images, both NeRF-based~\cite{rematas2022cvpr, yang2023cvpr-unisim} and 3DGS-based~\cite{zhou2024cvpr-drivinggs, yan2024eccv-streetgs, zhao2024eccv-tclcgs, fischer2024neurips-dgf, chen2024arxiv-omnire, hess2024arxiv-splatad} approaches have demonstrated high-fidelity offline radiance field reconstruction, enabling closed-loop autonomous driving simulation~\cite{yang2023cvpr-unisim, chen2024arxiv-omnire}.
However, radiance field mapping for even larger scenes at ground level with thousands of images remains challenging due to scene complexity and memory constraints.
\mbox{BlockNeRF}~\cite{tancik2022cvpr-blocknerf} addresses this by dividing scenes into overlapping blocks, training separate NeRFs per block, and consolidating them during rendering. 
Similarly, SiLVR~\cite{tao2024icra-silvr} employs a submap strategy for scalable NeRF mapping.
For 3DGS, hierarchical 3DGS~\cite{kerbl2024tog-hierarchical3dgs} introduces a level-of-detail hierarchy that enables real-time rendering of city-scale scenes. 
%
The aforementioned methods require time-consuming structure-from-motion preprocessing and offline divide-and-conquer processing, limiting their applicability for online missions. 

While there are several works on incremental mapping and SLAM with NeRF~\cite{sucar2021iccv, ortiz2022rss, sandstrom2023iccv-pointslam} or 3DGS~\cite{matsuki2024cvpr-monogs,keetha2024cvpr-splatam,zhu2025threedv-loopsplat,wei2024ral-gsfusion}, they primarily focus on bounded indoor scenes and struggle with our target scenarios.
Our proposed system enables incremental radiance and distance field mapping at the scale of previous offline methods~\cite{tancik2022cvpr-blocknerf, kerbl2024tog-hierarchical3dgs}, while achieving globally consistent 3D reconstruction through loop closure correction.

\section{Our Approach} 
\label{sec:main}


\begin{figure*}[h]
  \centering  
  \includegraphics[width=0.97\linewidth]{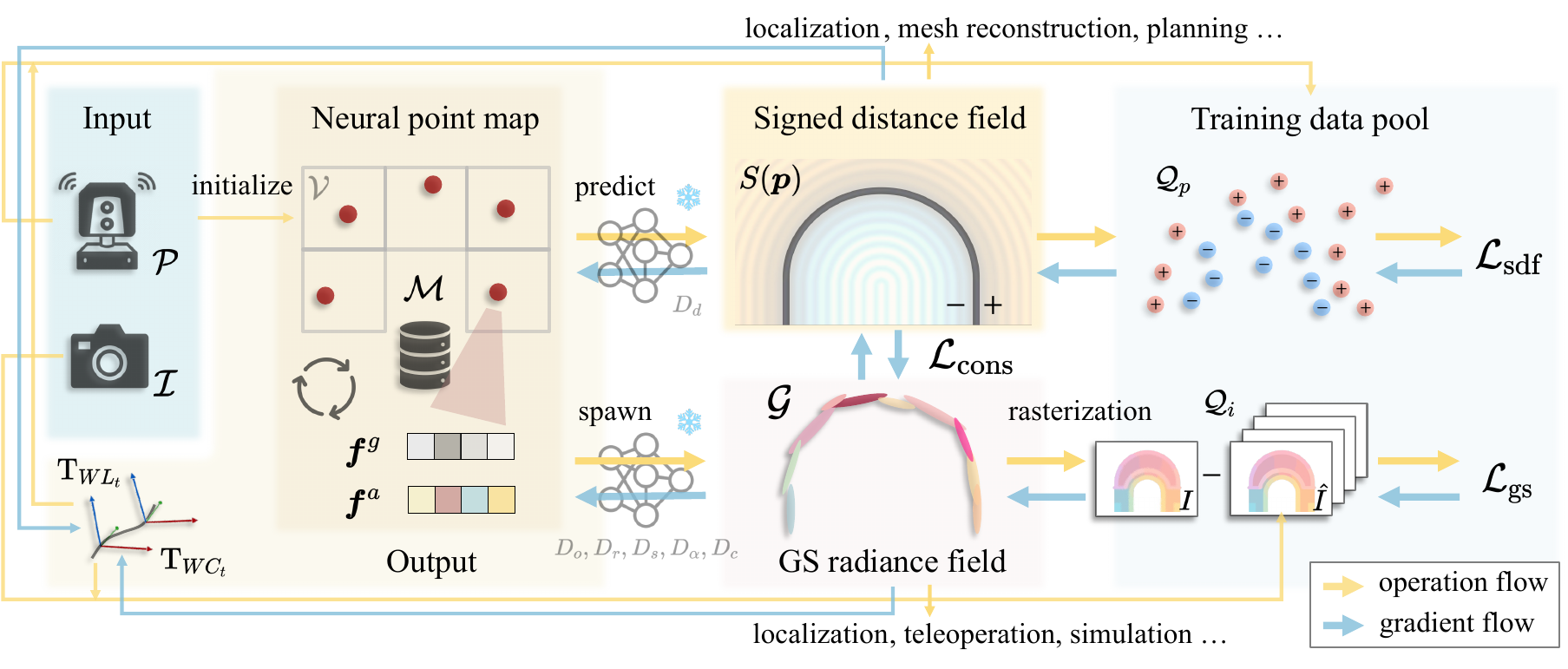}
  \setlength{\abovecaptionskip}{1pt}
  \caption{
  Overview of PINGS: We take a stream of LiDAR point clouds $\mathcal{P}$ and camera images $\mathcal{I}$ as input. 
  We initialize a neural point map $\mathcal{M}$ from $\mathcal{P}$ and maintain a training pool of SDF-labeled points $\mathcal{Q}_p$ and recent images $\mathcal{Q}_i$. 
  The map uses a voxel hashing structure $\mathcal{V}$ where each neural point stores geometric features $\vv{f}^g$ and appearance features $\vv{f}^a$. 
  These features are used to predict SDF values $S(\vv{p})$ at an arbitrary position $\vv{p}$ and spawn Gaussian primitives $\mathcal{G}$ through MLP decoders. 
  We compute three kind of losses: (1) Gaussian splatting loss $\mathcal{L}_{\text{gs}}$ comparing rendered images through differentiable rasterization and reference images in the training pool, (2) SDF loss $\mathcal{L}_{\text{sdf}}$ comparing predicted SDF and labels of the sampled points in the training pool, and (3) consistency loss $\mathcal{L}_{\text{cons}}$ to align the geometry of both representations. 
  The losses are backpropagated to optimize the neural point features $\vv{f}^g$ and $\vv{f}^a$.
  Meanwhile, we estimate LiDAR odometry by aligning the point cloud to current SDF and backpropagate $\mathcal{L}_{\text{gs}}$ to refine the camera poses.
  The final outputs are LiDAR poses $\mq{T}_{WL_t}$, camera poses $\mq{T}_{WC_t}$, and a compact neural point map $\mathcal{M}$ representing both SDF and Gaussian splatting radiance fields, enabling various robotic applications.  }
  \label{fig:overview}
  \vspace{-12pt}
\end{figure*}

Our approach, called PINGS, is a LiDAR-visual SLAM system that jointly builds globally consistent Gaussian splatting radiance fields and distance fields for large-scale scenes.

\textbf{Notation.}
In the following, we denote the transformation from coordinate frame $A$ to frame $B$ as $\mq{T}_{BA} \in \mathrm{SE}(3)$, such that \mbox{point $\vv{p}_B = \mq{T}_{BA} \vv{p}_A$}, with \mbox{rotation $\m{R}_{BA} \in \mathrm{SO}(3)$} and \mbox{translation $\vv{t}_{BA} \in \RR^{3}$}, where the rotation is also parameterized by a unit quaternion $\vv{q}$.
At timestep $t$, each sensor \mbox{frame $S_t$} (LiDAR frame $L_t$ or camera frame $C_t$) is related to the world frame $W$ by pose $\mq{T}_{WS_t}$, with $\mq{T}_{WS_0}$ fixed as identity.
%
%
We denote the rotation of a vector $\vv{v} \in \RR^3$ by a \mbox{quaternion $\vv{q}$} as $\vv{q}\vv{v}\vv{q}^{-1}$ and the multiplication of two quaternions as $\vv{q}_1\vv{q}_2$.




\textbf{Overview.}
%
We assume the robot is equipped with a LiDAR sensor and one or multiple  cameras.
At each \mbox{timestep $t$}, the input to our system is a LiDAR point cloud $\mathcal{P} = \{\vv{p} \in \RR^3\}$ and $M$ camera images $\mathcal{I} = \{\hat{\m{I}_i} \in \RR^{H \times W \times 3}\mid i=1, \ldots, M\}$ collected by the robot.
We assume the calibration of the LiDAR and cameras to be known but allow for the imperfect synchronization among the sensors.
Our system aims to simultaneously estimate the LiDAR pose $\mq{T}_{WL_t}$ while updating a point-based implicit neural map $\mathcal{M}$, which models both a SDF and a radiance field, as summarized in \figref{fig:overview}.
%

\subsection{Point-based Implicit Neural Map Representation}
\label{sec:pin_map}

We define our point-based implicit neural map $\mathcal{M}$ as a set of neural points, given by: 
\begin{equation}
  \mathcal{M}= \{\vv{m}_i = \left(\vv{x}_i, \vv{q}_i, \vv{f}_i^g, \vv{f}_i^a, \tau_i^c, \tau_i^u\right) \mid i=1, \ldots, N\}, 
\end{equation}
where each neural point $\vv{m}_i$ is defined in the world frame $W$ by a position $\vv{x}_i \in \mathbb{R}^3$ and a quaternion $\vv{q}_i \in \mathbb{R}^4$ representing the orientation of its own coordinate frame. 
Each neural point stores the optimizable geometric feature vector $\vv{f}_i^g \in \mathbb{R}^{F_g}$ and appearance feature vector $\vv{f}_i^a \in \mathbb{R}^{F_a}$. 
%
In addition, we keep track of each neural point's creation timestep $\tau_i^c$ and last update timestep $\tau_i^u$ to determine its active status and associate the neural point with the LiDAR pose $\mq{T}_{WL_{\tau}}$ at the middle \mbox{timestep  $\tau_i=\lfloor (\tau_i^c+\tau_i^u)/2 \rfloor$} between $\tau_i^c$ and $\tau_i^u$, thus allowing direct map manipulation through pose updates.

We maintain a voxel hashing~\cite{niessner2013tog} data structure $\mathcal{V}$ with a voxel resolution $v_p$ for fast neural point indexing and neighbor search, where each voxel stores at most one active neural point.
%
%

During incremental mapping, we dynamically update the neural point map based on point cloud measurements.
For each newly measured point $\vv{p}_W$ in the world frame, we check its corresponding voxel in $\mathcal{V}$. 
If no active neural point exists in that voxel, we initialize a new neural point $\vv{m}$ with the position $\vv{x} = \vv{p}_W$, an identity quaternion $\vv{q}=\left(1,0,0,0\right)$, and the feature vectors $\vv{f}^{g}=\vv{0},~\vv{f}^{a}=\vv{0}$.
Additionally, we define a local map $\mathcal{M}_l$ centered at the current LiDAR \mbox{position $\vv{t}_{WL_t}$}, which contains all active neural points within \mbox{radius $r_l$}.
To avoid incorporating inconsistent historical observations caused by odometry drift, both map optimization and odometry estimation are operated only within this local map $\mathcal{M}_l$.
After the map optimization at each timestep, we reassign the local map $\mathcal{M}_l$ into the global map $\mathcal{M}$.

Next, we describe how the neural points map $\mathcal{M}$ models both the SDF (\secref{subsec:nsdf}) and the radiance field (\secref{subsec:ngsf}).

\subsection{Neural Signed Distance Field}
\label{subsec:nsdf}
For the modeling and online training of a continuous SDF using the neural points, we follow the same strategy as in PIN-SLAM~\cite{pan2024tro} and present a recap in this section.

We model the SDF value $s$ at a query position $\vv{p}$ in the world frame $W$ conditioned on its nearby neural points.
For each neural point $\vv{m}_j$ in the k-nearest neighborhood $\mathcal{N}_p$ of $\vv{p}$, we define the relative coordinate $\vv{d}_j = \vv{q}_j (\vv{p} - \vv{x}_j)  \vv{q}_{j}^{-1}$ \mbox{denoting $\vv{p}$} in the local coordinate system of $\vv{m}_j$. 
%
Then, we feed the geometric feature vector $\vv{f}_j^g$ and the relative coordinate $\vv{d}_j$ to a globally shared SDF decoder $D_d$ to predict the SDF $s_j$:
\begin{align}
s_j = D_d(\vv{f}_j^g, \vv{d}_j).
\label{equ:sdf_prediction}
\end{align}

As shown in \figref{fig:explain}\,(a), the predicted SDF values $s_j$ of the neighboring neural points at the query position $\vv{p}$ are then interpolated as the final prediction $s = S(\vv{p})$, given by:
\begin{align}
  S(\vv{p}) =\sum_{j \in \mathcal{N}_p} \frac{w_j}{\sum_{k \in \mathcal{N}_p} w_k} s_j, 
\label{equ:interpolation}
\end{align}
with the interpolation weights $w_j={\left\|\vv{p} - \vv{x}_j\right\|^{-2}}$.

To optimize the neural SDF represented by the neural point geometric features $\{\vv{f}_i^g\}_{i=1}^{N}$ and the SDF decoder $D_d$, we sample points along the LiDAR rays around the measured end points and in the free space.
We take the projective signed distance along the ray as a pseudo SDF label for each sample point.
For incremental learning, we maintain a training data pool $\mathcal{Q}_p$ containing sampled points from recent scans, with a maximum capacity and bounded by a distance threshold from the current robot position.
At each timestep, we sample from the training data pool in batches and predict the SDF value at the sample positions.
The SDF training loss $\mathcal{L}_{\text{sdf}}$ is formulated as a weighted sum of the binary cross entropy loss term $\mathcal{L}_\text{bce}$ and the Eikonal loss term $\mathcal{L}_\text{eik}$, given by:
\begin{equation}
  \mathcal{L}_{\text{sdf}} = \lambda_{\text{bce}} \mathcal{L}_\text{bce} + \lambda_{\text{eik}} \mathcal{L}_\text{eik}.   
  \label{equ:sdf_loss}
\end{equation} 

The loss term $\mathcal{L}_\text{bce}$ applies a soft supervision on the SDF values by comparing the sigmoid activation of both the predictions and the pseudo labels.
The Eikonal loss term $\mathcal{L}_\text{eik}$ regularizes the SDF gradients by enforcing the Eikonal constraint~\cite{gropp2020icml}, which requires unit-length gradients $\|\nabla S(\vv{x})\|=1$ for the sampled points.
For more details regarding the SDF training, we refer readers to Pan~\etal~\cite{pan2024tro}.

The incrementally built neural SDF map can then be used for LiDAR odometry estimation and surface mesh extraction. 


\subsection{Neural Gaussian Splatting Radiance Field}
\label{subsec:ngsf}

\begin{figure*}[h]
  \centering  
  \includegraphics[width=0.97\linewidth]{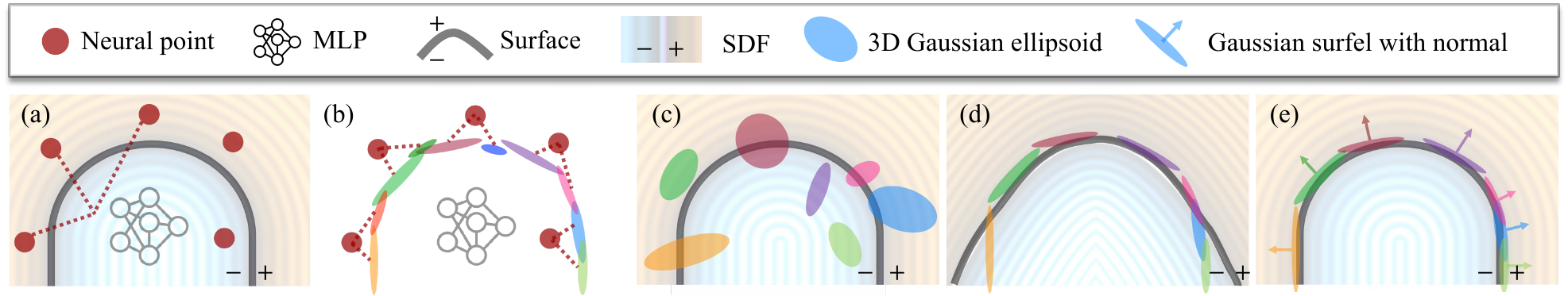}
  \setlength{\abovecaptionskip}{2pt}
  \caption{
  Example of neural point-based SDF prediction, Gaussian primitives spawning, and the geometric consistency of PINGS: 
  (a) SDF prediction at a query point through weighted interpolation of predictions from neighboring neural points.
  (b) Neural points spawning multiple Gaussian primitives to compose the radiance field.
  (c) Example of an accurate SDF but geometrically inaccurate radiance field with 3D Gaussian ellipsoids in regions with dense LiDAR coverage but sparse camera views, weak texture, or poor lighting.
  (d) Example of a geometrically accurate radiance field but inaccurate SDF in regions with rich visual data but sparse LiDAR measurements.
  (e) Our solution: flattening 3D Gaussian ellipsoids to surfels and enforcing geometric consistency by aligning surfel centers with the SDF zero-level set and aligning surfel normals with SDF gradients, resulting in accurate geometry for both fields.
  }
  \label{fig:explain}
  \vspace{-12pt}
\end{figure*}

We use camera image streams $\mathcal{I}$ to construct a radiance field by spawning Gaussian primitives from our neural point \mbox{map $\mathcal{M}$} and optimizing $\mathcal{M}$ via differentiable rasterization.

\textbf{Neural Point-based Gaussian Spawning.}
Inspired by Scaffold-GS~\cite{lu2024cvpr-scaffoldgs}, we use our neural points as anchor points for spawning Gaussian primitives, see \figref{fig:explain}\,(b).
For each neural point $\vv{m}$ lying within the current camera frustum, we \mbox{spawn $K$} Gaussian primitives by feeding its feature vectors ($\vv{f}^g$, $\vv{f}^a$) through globally shared MLP decoders.
We parameterize each spawned Gaussian primitive $\vv{g}$ with its \mbox{position $\vv{\mu} \in \mathbb{R}^{3}$} in the world frame, \mbox{rotation $\vv{r} \in \mathbb{R}^{4}$} in the form of a unit quaternion, \mbox{scale $\vv{s} \in \mathbb{R}^{3}$}, \mbox{opacity $\alpha \in \left[-1\,,1\right]$}, and RGB color $\vv{c} \in \left[0\,,1\right]^{3}$.

Each neural point spawns Gaussian primitives in its local coordinate frame defined by its position $\vv{x}$ and orientation $\vv{q}$. 
The world-frame position $\vv{\mu}_i$ of each spawned primitive is:
\begin{align}
\{\vv{\mu}_i = \vv{q} \vv{o}_i \vv{q}^{-1} + \vv{x} \mid \vv{o}_i \in  D_o(\vv{f}^g)\}_{i=1}^{K}\,,
\label{equ:gaussian_offset}
\end{align}
where $D_o$ is the offset decoder that maps the geometric \mbox{feature $\vv{f}^g$} to a set of $K$ local offsets $\{\vv{o}_i\}_{i=1}^{K}$, which are then transformed into the world frame through quaternion rotation and translation.
Likewise, the rotation $\vv{r}_i$ of each spawned Gaussian primitive is predicted by the rotation decoder $D_r$ and then rotated by quaternion $\vv{q}$ as:
\begin{align}
\label{equ:gaussian_rotation}
\{\vv{r}_i = \vv{q} \hat{\vv{r}}_i \mid \hat{\vv{r}}_i \in  D_r(\vv{f}^g)\}_{i=1}^{K}\,.
\end{align}

The scale decoder $D_s$ predicts each primitive's scale $\vv{s}_i$ as:
\begin{align}
\{\vv{s}_i\}_{i=1}^{K} & = D_s(\vv{f}^g)\,.
\label{equ:gaussian_scale}
\end{align}

We predict opacity values $\alpha$ in the range $[-1, 1]$ and treat only Gaussian primitives with positive opacity as being valid.
To adaptively control spatial density of Gaussian primitives based on viewing distance, we feed the geometric feature $\vv{f}^g$ and the view distance $\delta_v = \left\|\vv{x} - \vv{t}_{WC}\right\|_{2}$ into the opacity \mbox{decoder $D_{\alpha}$}.
This implicitly encourages the network to predict fewer valid Gaussians for distant points and more for nearby points, reducing computational load. 
The opacity value $\alpha_i$ for each Gaussian primitive is predicted as:
\begin{align}
\{\alpha_i\}_{i=1}^{K} & = D_{\alpha}(\vv{f}^g, \delta_v)\,. 
\label{equ:gaussian_opacity}
\end{align}

For view-dependent color prediction, we take a different approach than the spherical harmonics used in 3DGS~\cite{kerbl2023tog-3dgs}.
We feed the appearance feature $\vv{f}^a$ and the \mbox{view direction $\vv{d}_v = (\vv{x} - \vv{t}_{WC})/\delta_v$} to the color decoder $D_{c}$ to predict the color $\vv{c}_i$ of each Gaussian primitive, given by:
\begin{align}
  \{\vv{c}_i\}_{i=1}^{K} & = D_{c}(\vv{f}^a, \vv{q}^{-1} \vv{d}_v \vv{q})\,,
\label{equ:gaussian_color}
\end{align}
where the view direction $\vv{d}_v$ is also transformed into the local coordinate system of the neural point.

Note that we treat position, rotation, scale, and opacity as geometric attributes of a Gaussian primitive, using the geometric feature $\vv{f}^g$ for their prediction, while using the appearance feature $\vv{f}^a$ to predict color.

\textbf{Gaussian Splatting Rasterization.}
We gather all the valid Gaussians primitives $\mathcal{G}$~\cite{kerbl2023tog-3dgs} spawned at the current viewpoint:
\begin{align}
  \mathcal{G}= \{\vv{g}_i = \left(\vv{\mu}_i\,,\vv{r}_i\,,\vv{s}_i,\,\vv{c}_i\,,\alpha_i \right) \mid i=1, \ldots, N_{g}\}\,. 
\end{align}

The distribution of each Gaussian primitive $\vv{g}_i$ in the world frame is represented as: 
\begin{align}
\mathcal{N}(\vv{x}; \vv{\mu}_i, \vv{\Sigma}_i) = \mathrm{exp}\left(-\frac{1}{2}(\vv{x} - \vv{\mu}_i)^\top \vv{\Sigma}_i^{-1} (\vv{x} - \vv{\mu}_i)\right)\,,
\end{align}
where the covariance matrix $\vv{\Sigma}_i$ is reparameterized as:
\begin{align}
\vv{\Sigma}_i = \m{R}(\vv{r}_i) \, \m{S}(\vv{s}_i) \, \m{S}(\vv{s}_i)^\top \, \m{R}(\vv{r}_i)^\top\,, 
\end{align}
where $\m{R}(\vv{r}_i) \in \mathit{SO}(3)$ is the rotation matrix derived from the quaternion $\vv{r}_i$ and $\m{S}(\vv{s}_i) = \mathrm{diag}(\vv{s}_i) \in \mathbb{R}^{3\times 3}$ is the diagonal scale matrix composed of the \mbox{scale $\vv{s}_i$} on each axis. 
Using a tile-based rasterizer~\cite{zwicker2001pv}, we project the Gaussian primitives to the 2D image plane and sort them according to depth efficiently. The projected Gaussian distribution is:
\begin{align}
  \vv{\mu}^{\prime} = \pi(\mq{T}_{CW} \vv{\mu})\,,  ~~~  \vv{\Sigma^{\prime}} = \m{J} \m{W} \vv{\Sigma} \m{W}^\top \m{J}^\top\,,
\end{align}
where $\vv{\mu}^{\prime}$ and $\vv{\Sigma^{\prime}}$ are the projected mean and covariance, \mbox{$\pi$ denotes} the perspective projection, $\m{J}$ is the Jacobian of the projective transformation, and $\m{W}$ is the viewing transformation deduced from current camera pose $\mq{T}_{WC}$.
The rendered RGB image $\m{I}$ at each pixel $\vv{u}$ is computed via alpha blending:
\begin{align}
  \m{I}(\vv{u}) &= \sum_{i \in \mathcal{G}(\vv{u})} w_i \vv{c}_i\,,
\end{align}
where the weight $w_i$ of each of the depth-sorted Gaussian primitives $\mathcal{G}(\vv{u})$ covering pixel $\vv{u}$ is given by:
\begin{align}
w_i = T_i \sigma_i\,, ~ T_i &= \prod_{j=1}^{i-1} (1 - \sigma_j)\,,~  \sigma_i = \mathcal{N} (\vv{u}; \vv{\mu}^{\prime}_i, \vv{\Sigma}^{\prime}_i ) \alpha_i ,
\label{equ:gaussian_weight}
\end{align}
%
where $\sigma_i$ is the projected opacity of the $i$-th Gaussian primitive, computed using the 2D Gaussian density function $\mathcal{N} (\vv{u}; \vv{\mu}^{\prime}_i, \vv{\Sigma}^{\prime}_i )$ evaluated at pixel $\vv{u}$ with the projected mean $\vv{\mu}^{\prime}_i$ and covariance $\vv{\Sigma}^{\prime}_i$.

\textbf{Gaussian Surfels Training.}
To achieve accurate and multi-view consistent geometry, we adopt Gaussian Surfels~\cite{dai2024siggraph-gaussian-surfels}, a state-of-the-art 2DGS representation~\cite{huang2024siggraph-2dgs}, by flattening 3D Gaussian ellipsoids into 2D disks (last dimension of scale $s^z=0$).
For each \mbox{pixel $\vv{u}$}, we compute the surfel depth $d\left(\vv{u}\right)$ as the ray-disk intersection distance, and obtain the \mbox{normal $\vv{n}$} as the third column of the rotation matrix $\m{R}(\vv{r})$.
%
Using alpha blending, we render the depth \mbox{map $\m{D}$} and the normal \mbox{map $\m{N}$} using the weights $w_i$ calculated in \eqref{equ:gaussian_weight}:
\begin{align}
  \m{D}(\vv{u}) = \sum_{i \in \mathcal{G}(\vv{u})} w_i d_i\left(\vv{u}\right)\,, ~~~\m{N}(\vv{u}) = \sum_{i \in \mathcal{G}(\vv{u})} w_i \vv{n}_i\,.
\end{align}
%

Given the training view with the RGB image $\widehat{\m{I}}$ and the sparse depth map $\widehat{\m{D}}$ projected from the LiDAR point cloud, we define the Gaussian splatting loss $\mathcal{L}_{\text{gs}}$ combining the photometric rendering $\mathcal{L}_{\text{photo}}$, depth \mbox{rendering $\mathcal{L}_{\text{depth}}$}, and area regularization $\mathcal{L}_{\text{area}}$ terms, given by:
\begin{align}
\label{equ:gaussian_splatting_loss}
\mathcal{L}_{\text{gs}} & = \lambda_{\text{photo}} \mathcal{L}_{\text{photo}} + \lambda_{\text{depth}}\mathcal{L}_{\text{depth}} + \lambda_{\text{area}}\mathcal{L}_{\text{area}}, \\ 
\mathcal{L}_{\text{photo}} & = 0.8 \cdot L_{1}\left(\m{I},\widehat{\m{I}}\right) + 0.2 \cdot L_{\text{ssim}}\left(\m{I},\widehat{\m{I}}\right), \\
\mathcal{L}_{\text{depth}} & = L_{1}\left(\m{D},\widehat{\m{D}}\right)\,, \\
\mathcal{L}_{\text{area}} & = \sum_{\vv{g}_i \in \mathcal{G}} s^x_i \cdot s^y_i\,,
\end{align}
where $L_{1}$ is the L1 loss, $L_{\text{ssim}}$ is the structural similarity index measure (SSIM) loss~\cite{wang2004tip}, $s^x_i$ and $s^y_i$ are the scales of the Gaussian surfel $\vv{g}_i$. The area loss term $\mathcal{L}_{\text{area}}$ encourages minimal overlap among the surfels covering the surface. 

%
To handle inaccurate camera poses resulting from imperfect LiDAR odometry and camera-LiDAR synchronization, we jointly optimize the camera poses on a manifold during radiance field training~\cite{matsuki2024cvpr-monogs}.
%
We also account for real-world lighting variations by optimizing per-frame exposure parameters~\cite{kerbl2024tog-hierarchical3dgs}.

\subsection{Joint Optimization with Geometric Consistency}
\label{subsec:cons}
To enforce mutual alignment between the surfaces represented by the SDF and Gaussian splatting radiance field, we futhermore propose to jointly optimize the geometric consistency.
This joint optimization helps resolve geometric ambiguities in the radiance field through the direct surface description of SDF, while simultaneously refining SDF's accuracy in regions with sparse LiDAR measurements using the dense photometric cues and multi-view consistency from the radiance field, see \figref{fig:explain}\,(c), (d), and (e).
For each sampled Gaussian surfel, we randomly sample points along its normal direction $\vv{n}$ from the center $\vv{\mu}$ with random offsets $\epsilon \sim U\left(-\epsilon_{\text{max}},\epsilon_{\text{max}}\right)$.
We enforce geometric consistency between the SDF and Gaussian surfels through a two-part consistency loss $\mathcal{L}_{\text{cons}}$, given by:
\begin{align}
  \label{equ:geometric_consistency_loss}
  \mathcal{L}_{\text{cons}} & = \lambda_{\text{cons}}^{\text{d}} \mathcal{L}_{\text{cons}}^{\text{d}} + \lambda_{\text{cons}}^{\text{v}} \mathcal{L}_{\text{cons}}^{\text{v}} , \\
  \mathcal{L}_{\text{cons}}^{\text{d}} & = \sum_{\vv{g}_i \in \mathcal{G}}  \left|S\left(\vv{\mu}_i+\epsilon_i\vv{n}_i \right) - \epsilon_i\right| , \\
  \mathcal{L}_{\text{cons}}^{\text{v}} & = \sum_{\vv{g}_i \in \mathcal{G}} \left(1 - \frac{\nabla S\left(\vv{\mu}_i+\epsilon_i\vv{n}_i\right)^{\tr} \vv{n}_i}{\norm{\nabla S\left(\vv{\mu}_i+\epsilon_i\vv{n}_i\right)}} \right) ,
\end{align}
where $\mathcal{L}_{\text{cons}}^{\text{d}}$ enforces SDF values to match sampled offsets via an L1 loss, and $\mathcal{L}_{\text{cons}}^{\text{v}}$ aligns SDF gradients $\nabla S$ at the sampled points with surfel normals $\vv{n}$ using cosine distance. 

We define the total loss $\mathcal{L}$ given by the sum of the SDF loss $\mathcal{L}_{\text{sdf}}$ in \eqref{equ:sdf_loss}, Gaussian splatting loss $\mathcal{L}_{\text{gs}}$ in \eqref{equ:gaussian_splatting_loss}, and the geometric consistency loss $\mathcal{L}_{\text{cons}}$ in \eqref{equ:geometric_consistency_loss}:
\begin{align}
\mathcal{L} = \mathcal{L}_{\text{sdf}} + \mathcal{L}_{\text{gs}} + \mathcal{L}_{\text{cons}}.
\end{align}

We jointly optimize the neural point features $\left\{\vv{f}^g_i, \vv{f}^a_i\right\}_{i=1}^{N}$, decoder parameters, camera poses, and exposure correction parameters to minimize the total loss $\mathcal{L}$.


\subsection{PINGS LiDAR-Visual SLAM System}
We devise a LiDAR-visual SLAM system called PINGS using the proposed map representation, built on top of the LiDAR-only PIN-SLAM~\cite{pan2024tro} system. 
PINGS alternates between two main steps:
\mbox{(i) mapping:} incremental learning of the local neural point map $\mathcal{M}_l$, which jointly models the SDF and Gaussian splatting radiance field, and
\mbox{(ii) localization:} odometry estimation using the learned SDF.
In addition, loop closure detection and pose graph optimization run in parallel.

%

%
%
We initialize PINGS with 600 iterations of SDF training using only the first LiDAR scan.
At subsequent timesteps, we jointly train the SDF and radiance field for 100 iterations.
To prevent catastrophic forgetting during incremental mapping, we freeze decoder parameters after 30 timesteps and only update neural point features.
We found the decoders converge on learning the interpretation capability within these 30 frames. 

We maintain sliding window-like training pools $\mathcal{Q}_p$ and $\mathcal{Q}_i$ containing SDF-labeled sample points and image data whose view frustum overlaps with the local map $\mathcal{M}_l$, respectively.
Each training iteration samples one image from $\mathcal{Q}_i$ and 8192 points from $\mathcal{Q}_p$ for optimization.

We estimate LiDAR odometry by aligning each new scan to the SDF's zero level set using an efficient Gauss-Newton optimization~\cite{wiesmann2023ral-locndf} that requires only SDF values and gradients queried at source point locations, eliminating the need for explicit point correspondences.
%
%
Initial camera poses are derived from the LiDAR odometry and extrinsic calibration, then refined via gradient descent during the radiance field optimization to account for imperfect camera-LiDAR synchronization, as described in \secref{subsec:ngsf}.

In line with PIN-SLAM, we detect loop closures using the layout and features of the local neural point map. 
We then conduct pose graph optimization to correct the drift of the LiDAR odometry and get globally consistent poses.
We move the neural points along with their associated LiDAR frames to keep a globally consistent map.
Suppose $\Delta \mq{T}$ is the pose correction matrix of LiDAR frame $L_{i}$ after pose graph optimization, we update the position $\vv{x}$ and orientation $\vv{q}$ of each neural point associated with $L_{i}$ as:
\begin{align} 
  \vv{x} \leftarrow \Delta \mq{T} \vv{x}\,, ~~~
  \vv{q} \leftarrow \Delta \vv{q} \vv{q}\,,
\end{align}
where $\Delta \vv{q}$ is the rotation part of $\Delta \mq{T}$ in the form of a quaternion. 
Since the positions, rotations, and colors of the spawned Gaussian primitives are predicted in the local frames of their anchor neural points, see \eqref{equ:gaussian_offset}, \eqref{equ:gaussian_rotation}, and \eqref{equ:gaussian_color}, they automatically transform with their anchor neural points, thus maintaining the global consistency of the radiance field.


%
PINGS aims to build static distance and radiance fields without artifacts from dynamic objects.
Since measured points with large SDF values in stable free space likely correspond to dynamic objects~\cite{schmid2023ral-dynablox}, we identify neural points representing dynamic objects through SDF thresholding. 
We disable Gaussian primitive spawning for these points, effectively preventing dynamic objects from being rendered from the radiance field.

\section{Experimental Evaluation}
\label{sec:exp}

%
The main focus of this paper is an approach for LiDAR-visual SLAM that unifies Gaussian splatting radiance fields and signed distance fields by leveraging their mutual consistency within a point-based implicit neural map representation.

%
We present our experiments to show the capabilities of our method called PINGS. 
The results of our experiments support our key claims, which are:
(i) PINGS achieves better RGB and geometric rendering at novel views by constraining the Gaussian splatting radiance field using the SDF;
(ii) PINGS builds a more accurate SDF for more accurate localization and surface reconstruction by leveraging dense photometric cues from the radiance field; 
(iii) PINGS enables large-scale globally consistent mapping with loop closures;
(iv) PINGS builds a more compact map than previous methods for both radiance and distance fields.

\subsection{Experimental Setup}

\subsubsection{Datasets}

We evaluate PINGS on self-collected in-house car datasets and the Oxford Spires dataset~\cite{tao2024arxiv-oxford}. 
Our in-house car datasets were collected using a robot car equipped with four Basler Ace cameras providing \SI{360}{\degree} visual coverage and an Ouster OS1-128 LiDAR (\SI{45}{\degree} vertical FOV, 128 beams) mounted horizontally, both operating at \SI{10}{\hertz}.
We calibrate the LiDAR-camera system using the method proposed by Wiesmann~\etalcite{wiesmann2024ral} and generate reference poses through offline LiDAR bundle adjustment~\cite{wiesmann2024arxiv-ba}, incorporating RTK-GNSS data, point cloud alignment as well as constraints from precise geo-referenced terrestrial laser scans.

\begin{figure}[!t]
  \centering  
  \includegraphics[width=0.99\linewidth]{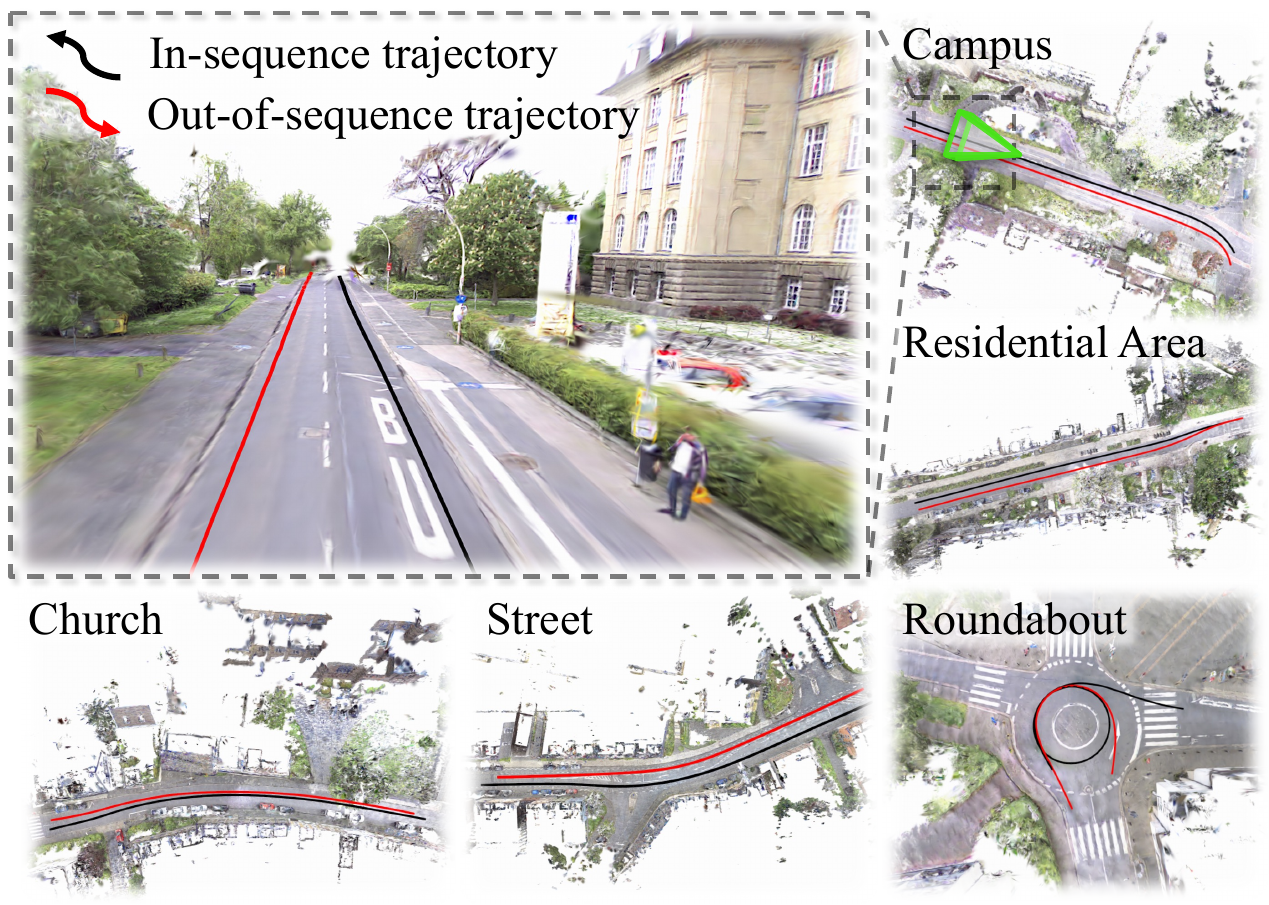}
  \setlength{\abovecaptionskip}{-9pt}
  \caption{
    Visualization of the five scenes from our \textit{in-house car dataset} used for novel view rendering evaluation. For each scene, we show a bird's eye view rendering from the radiance field built by PINGS, with a detailed zoom-in of the \textit{Campus} scene. The robot car traversed each area twice in opposite directions with lane-level lateral displacement. The black trajectory provided images for training (sampled) and in-sequence testing (unsampled), while the red trajectory provided out-of-sequence testing views.
    }
  \label{fig:car_dataset}
  \vspace{-14pt}
\end{figure}

We evaluate the SLAM localization accuracy and scalability of PINGS on two long sequences from our dataset: a \SI{5}{\km} sequence with around 10,000 LiDAR scans and 40,000 images, and a second sequence which is a bit shorter.
Both sequences traverse the same area in opposite directions on the same day.
For better quantitative evaluation of the radiance field mapping quality, we select five subsequences of 150 LiDAR scans and 600 images each, as shown in \figref{fig:car_dataset}. 
Having sequences captured in opposite driving directions and lane-level lateral displacement allows us to evaluate novel view rendering from substantially different viewpoints from the training views (out-of-sequence testing views), which is a critical capability for downstream tasks such as planning and simulation.

We evaluate surface reconstruction accuracy on the Oxford Spires dataset~\cite{tao2024arxiv-oxford}, which provides a millimeter-accurate reference map from a Leica RTC360 terrestrial laser scanner. 
The data was collected using a handheld system equipped with three global-shutter cameras and a 64-beam LiDAR.

\subsubsection{Parameters and Implementation Details}

For mapping parameters, we set the local map radius $r_l$ to \SI{80}{\meter}, voxel resolution $v_p$ to \SI{0.3}{\meter}, and maximum sample offset for consistency loss $\epsilon_{\text{max}}$ to $0.5\,v_p$.
The training data pool $\mathcal{Q}_d$ has a capacity of $2 \cdot 10^7$ SDF-labeled sample points, and $\mathcal{Q}_i$ has a capacity of $200$ images.
During map optimization, we use Adam~\cite{kingma2015iclr} with learning rates of $0.002$ for neural point features, $0.001$ for decoder, camera poses and exposure corrections parameters.
The neural point feature dimensions $F_g$ and $F_a$ are set to $32$ and $16$, respectively. 
All decoders use shallow MLPs with one hidden layer of $128$ neurons.
Each neural point spawns $K=8$ Gaussian primitives.
For decoder activations, we use sigmoid for SDF decoder $D_d$ and color decoder $D_c$, tanh for offset decoder $D_o$ and opacity decoder $D_{\alpha}$, and exponential for scale decoder $D_s$.
The Gaussian spawning offset is scaled to $[-2v_p, 2v_p]$, and the scale output is clamped to a maximum of $2v_p$.
The rotation decoder $D_r$ output is normalized to valid unit quaternions.
The weights for different loss terms are set to: $\lambda_{\text{bce}}=1.0$, $\lambda_{\text{eik}}=0.5$, $\lambda_{\text{photo}}=1.0$, $\lambda_{\text{depth}}=0.01$, $\lambda_{\text{area}}=0.001$, and $\lambda_{\text{cons}}^d=\lambda_{\text{cons}}^v=0.02$.

For training and testing, we use image resolutions of $512 \times 1,032$ for the in-house car dataset and $540 \times 720$ for the Oxford Spires dataset.
The experiments are carried out on a single NVIDIA A6000 GPU.

\subsection{Novel View Rendering Quality Evaluation}

\begin{table*}[!t]
	\caption{Quantitative comparison of rendering quality on the \textit{in-house car dataset}. We evaluate rendering photorealism using PSNR, SSIM, and LPIPS metrics, and geometric accuracy using Depth-L1 error (in \si{m}). Best results are shown in \textbf{bold}, second best are \underline{underscored}.}
	\centering
\resizebox{0.89\textwidth}{!}{
\begin{tabular}{c|c|cccc|cccc}
\toprule 
\multirow{2}{*}{Sequence}  & \multirow{2}{*}{Method} & \multicolumn{4}{c|}{In-Sequence Testing View} & \multicolumn{4}{c}{Out-of-Sequence Testing View} \\
& & PSNR$\uparrow$ & SSIM$\uparrow$ & LPIPS$\downarrow$ & Depth-L1$\downarrow$ & PSNR$\uparrow$ & SSIM$\uparrow$ & LPIPS$\downarrow$ & Depth-L1$\downarrow$ \\
\midrule 

\multirow{5}{*}{Church} 
& 3DGS                & 18.02	& 0.62 & 0.52 &	1.45 	  &    16.37 & 0.59	& 0.52 & 1.22  \\ 
& GSS                & 18.04	& 0.63 & 0.51	& 1.37 	  &    16.44 & 0.60	& 0.52 & 1.25	 \\ 
& Neural Point + 3DGS & 20.89	& 0.71	& 0.41	& 0.80  &  	 19.56	& 0.70	& 0.42	& 0.78	\\
& Neural Point + GGS & \underline{22.56}	& \underline{0.75} & \underline{0.36}	& \bf{0.43}    &	 \underline{20.48} & \underline{0.74}	& \underline{0.38} & \underline{0.47}  \\
& PINGS (Ours)        & \bf{22.93}	& \bf{0.78} & \bf{0.33}	& \bf{0.43}	  &    \bf{20.79} & \bf{0.76} & \bf{0.34} & \bf{0.46}	 \\
\midrule 
\multirow{5}{*}{Residential Area} 
& 3DGS                & 17.60	& 0.58 & 0.52	& 2.22     &  	14.63 & 0.53 &	0.56 & 2.78  \\ 
& GSS                & 17.56	& 0.59 & 0.51 &	2.26     &    14.80 & 0.54	& 0.55 & 2.68  \\ 
& Neural Point + 3DGS & 21.10	& 0.71	& 0.38	& 0.89   &    18.34	& 0.65	& 0.42	& 0.93 \\
& Neural Point + GSS & \underline{22.33}	& \underline{0.73} & \underline{0.35}	& \bf{0.53}  	 &     \underline{19.31}	& \underline{0.69}	& \underline{0.38} & \underline{0.69}	 \\
& PINGS (Ours)        & \bf{22.67}	& \bf{0.77} & \bf{0.30}	& \bf{0.53}     &  	\bf{19.48} & \bf{0.71}	& \bf{0.34} & \bf{0.68}  \\
\midrule 
\multirow{5}{*}{Street} 
& 3DGS                & 16.39 &	0.56 & 0.55 &	2.09     &    15.73 & 0.57 &	0.53 & 2.30 \\ 
& GSS                & 16.85 &	0.59 & 0.53 &	1.87     &    16.01 & 0.59 & 0.52 & 2.15 \\ 
& Neural Point + 3DGS & 19.74	& 0.68	& 0.42	& 0.81  &    18.02	& 0.64	& 0.44	& 0.79	\\
& Neural Point + GSS & \underline{22.13}	& \underline{0.75} & \underline{0.35}	& \underline{0.29}     &     \underline{19.09}	& \underline{0.69}	& \underline{0.40} & \underline{0.49} \\
& PINGS (Ours)        & \bf{22.45}	& \bf{0.78} & \bf{0.32}	& \bf{0.28}     &     \bf{19.34}	& \bf{0.71}	& \bf{0.37} & \bf{0.47} \\
\midrule 
\multirow{5}{*}{Campus} 
& 3DGS                & 17.38 &	0.57 & 0.52 &	2.70     &    14.88 & 0.49 &	0.58 & 3.60 \\ 
& GSS                & 17.34 &	0.59 & 0.51 &	2.36     &    14.96 & 0.51 &	0.57 & 3.42 \\ 
& Neural Point + 3DGS & 20.04	& 0.67 & 0.40  & 1.06   &   17.83	& 0.60	& 0.44	& 1.19 \\
& Neural Point + GSS & \underline{21.82}	& \underline{0.72} & \underline{0.35}	& \underline{0.65}    &  \underline{18.71}	& \underline{0.64}	& \underline{0.41}	& \bf{0.79} \\
& PINGS (Ours)        & \bf{22.40}	& \bf{0.76} & \bf{0.31}	& \bf{0.64}     &     \bf{18.91}	& \bf{0.66}	& \bf{0.38} & \underline{0.80} \\
\midrule 
\multirow{5}{*}{Roundabout} 
& 3DGS                &  21.20	& 0.71 & 0.39 &	0.87     &  18.97	& 0.69	& 0.40	& 0.85 \\ 
& GSS                &  21.74	& 0.72 & 0.38	& \underline{0.55}     &  19.10	& 0.69	& 0.40	& 0.60 \\ 
& Neural Point + 3DGS &  21.44	& \underline{0.75} & \underline{0.35}	& 0.72     & 19.23	& 0.72	& 0.37	& 0.78	\\
& Neural Point + GSS & \bf{23.54}	 & \bf{0.82} &	\bf{0.28} &	\bf{0.47}    &   \underline{20.22} & \bf{0.78}	& \underline{0.31}	& \underline{0.55}\\
& PINGS (Ours)        & \underline{23.45}	& \bf{0.82} & \bf{0.28}	& \bf{0.47}	   &  \bf{20.23}	& \underline{0.77}	& \bf{0.30} & \bf{0.54} \\
\bottomrule 
\end{tabular}
}
\vspace{-4pt}
\label{tab:rendering} 
\end{table*}

\begin{figure*}[htbp]
  \centering  
  \includegraphics[width=0.98\linewidth]{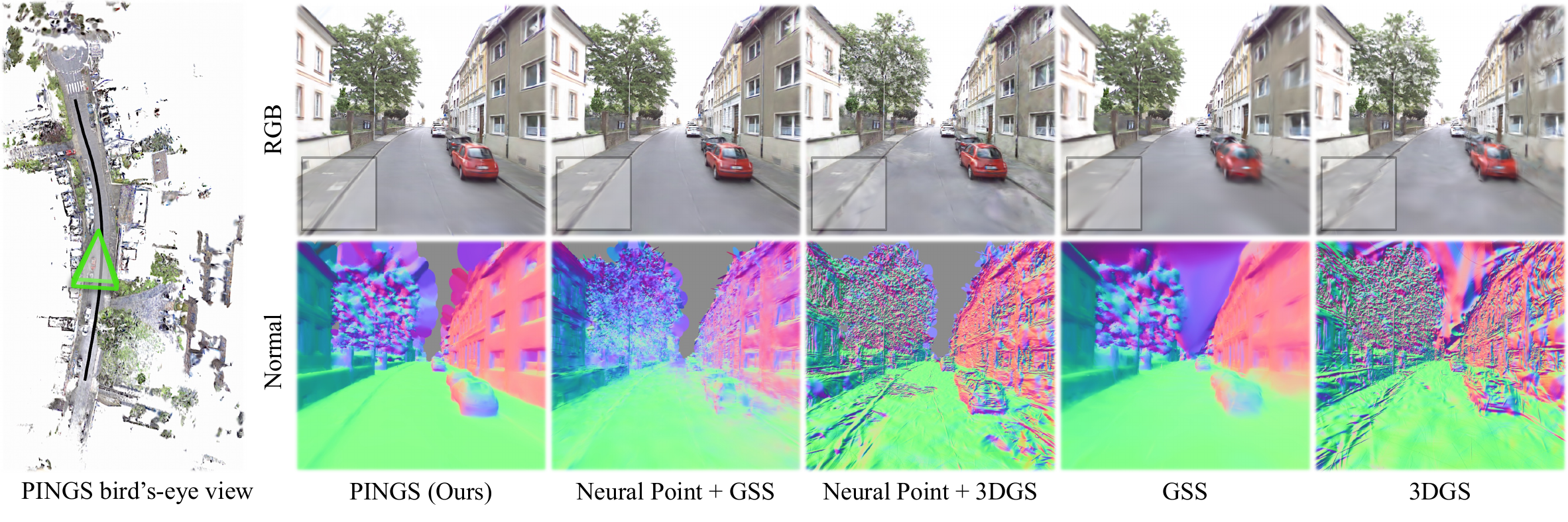}
  \setlength{\abovecaptionskip}{1pt}
  \caption{Qualitative comparison of rendering quality on the \textit{in-house car dataset}. Left: Bird's eye view rendering of the \textit{Church} scene, showing the training view trajectory (black line) and the test viewpoint for comparison (green camera frustum). Right: RGB and normal map renderings from different methods at the test viewpoint, with detailed comparison of curb and sidewalk rendering in the highlighted box.}

  \label{fig:qualitative_rendering_church_1}
  \vspace{-10pt}
\end{figure*}



\begin{table*}[!htbp]
  \caption{Quantitative evaluation of surface reconstruction quality on the \textit{Oxford-Spires dataset}. We use the metrics include accuracy error (in \si{m}), completeness error (in \si{m}), and Chamfer distance (in \si{m}), as well as precision, recall and F-score (with \SI{0.1}{\meter} threshold).  $\dag$ denotes methods requiring offline batch processing. Best results are shown in \textbf{bold}, second best are \underline{underscored}. }
  \vspace{-2pt}
  \label{tab:recon_table}
  \centering
  \resizebox{0.88\textwidth}{!}{
  \begin{tabular}{c|c|c c c|c c c}
      \specialrule{1.5pt}{1pt}{1pt}
      Sequence& Method &  Accuracy $\downarrow$ & Completeness $\downarrow$ & Chamfer Distance $\downarrow$ & Precision $\uparrow$ & Recall $\uparrow$ & F-score $\uparrow$ \\
      \toprule
      \multirow{6}{*}{Blenheim Palace 05} & OpenMVS$^\dag$ & 0.126 & 1.045 & 0.586 & 0.574 & 0.381 & 0.458 \\
      & Nerfacto$^\dag$ & 0.302 & 0.676 & 0.489 & 0.388 & 0.257 & 0.309 \\
      & GSS & 0.204 & 0.254 & 0.229 & 0.271 & 0.261 & 0.266 \\
      & VDB-Fusion & 0.098 & \bf{0.123} & 0.111 & 0.646 & \bf{0.746} & 0.692 \\
      & PIN-SLAM & \underline{0.078} & 0.136 & \underline{0.107} & \underline{0.768} & 0.712 & \underline{0.739} \\
      & PINGS (Ours) & \bf{0.072} & \underline{0.133} & \bf{0.102} & \bf{0.794} & \underline{0.726} & \bf{0.758} \\
      \midrule
      \multirow{6}{*}{Christ Church 02} & OpenMVS$^\dag$ & \bf{0.046} & 5.381 & 2.714 & \bf{0.886} & 0.266 & 0.410 \\
      & Nerfacto$^\dag$ & 0.219 & 4.435 & 2.327 & 0.532 & 0.254 & 0.343 \\
      & GSS  & 0.174 & 0.292 & 0.233 & 0.407 & 0.301 & 0.346 \\
      & VDB-Fusion & 0.098 & \bf{0.243} & 0.171 & 0.655 & \bf{0.524} & 0.582 \\
      & PIN-SLAM & 0.069 & 0.252 & \underline{0.160} & 0.812 & 0.497 & \underline{0.617} \\
      & PINGS (Ours) & \underline{0.067} & \underline{0.251} & \bf{0.159} & \underline{0.815} & \underline{0.502} & \bf{0.622} \\
      \midrule
      \multirow{6}{*}{Keble College 04} & OpenMVS$^\dag$  & \bf{0.067} & 0.342 & 0.205 & \bf{0.918} & 0.718 & \bf{0.806}  \\
      & Nerfacto$^\dag$  & 0.137 & 0.150 & 0.144 & 0.654 & 0.709 & 0.680 \\
      & GSS  & 0.171 & 0.162 & 0.167 & 0.424 & 0.518 & 0.466 \\
      & VDB-Fusion & 0.103 & \bf{0.101} & \underline{0.102} & 0.639 & \bf{0.821} & 0.719 \\
      & PIN-SLAM & 0.096 & 0.108 & \underline{0.102} & 0.701 & 0.793 & 0.744 \\
      & PINGS (Ours) & \underline{0.093} & \underline{0.106} & \bf{0.099} & \underline{0.705} & \underline{0.799} & \underline{0.749} \\
      \midrule
      \multirow{6}{*}{Observatory Quarter 01} & OpenMVS$^\dag$ & \bf{0.048} & 0.622 & 0.335 & \bf{0.902} & 0.618 & \bf{0.734} \\
      & Nerfacto$^\dag$ & 0.197 & 0.398 & 0.298 & 0.587 & 0.598 & 0.592\\
      & GSS & 0.179 & 0.184 & 0.181 & 0.377 & 0.443 & 0.407 \\
      & VDB-Fusion & 0.123 & \bf{0.109} & \underline{0.116} & 0.573 & \bf{0.737} & 0.645 \\
      & PIN-SLAM & 0.105 & 0.129 & 0.117 & 0.654 & 0.677 & 0.665 \\ 
      & PINGS (Ours) & \underline{0.102} & \underline{0.124} & \bf{0.113} & \underline{0.659} & \underline{0.705} & \underline{0.681} \\
      \specialrule{1.5pt}{1pt}{1pt}   
  \end{tabular}
  }
  \vspace{-10pt}
\end{table*}

\begin{figure*}[htbp]
  \centering  
  \includegraphics[width=0.92\linewidth]{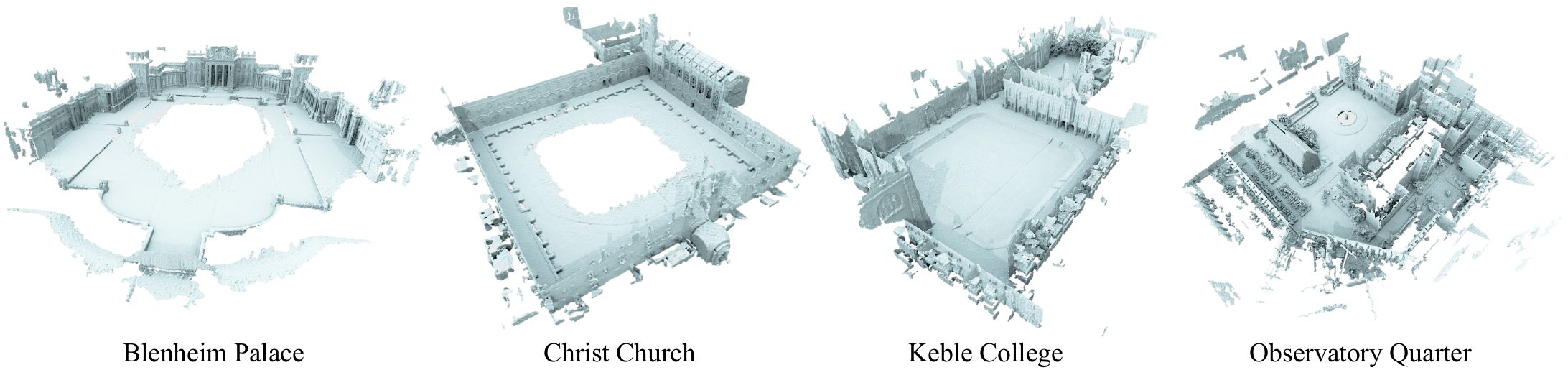}
  \setlength{\abovecaptionskip}{-1pt}
  \caption{Qualitative results of the surface mesh reconstruction by PINGS on the \textit{Oxford-Spires dataset}. The meshes are extracted using marching cubes algorithm from the SDF with a resolution of \SI{0.1}{\meter}.}
  \label{fig:qualitative_surface_reconstruction}
  \vspace{-12pt}
\end{figure*}

We evaluate novel view rendering quality on five subsequences from the in-house car dataset.
For quantitative evaluation, we employ standard metrics: PSNR~\cite{mildenhall2020eccv}, SSIM~\cite{wang2004tip}, and LPIPS~\cite{zhang2018cvpr-lpips} to assess photorealism, along with Depth-L1 error to measure geometric accuracy.
We compute these metrics for both in-sequence and out-of-sequence testing views.
We consider the following methods for comparison:
\begin{itemize}
\item \textit{3DGS}~\cite{kerbl2023tog-3dgs}: An incremental training variant of 3DGS initialized with LiDAR measurements and supervised with the depth rendering \mbox{loss $\mathcal{L}_{\text{depth}}$} as defined in \secref{subsec:ngsf}.

\item \textit{GSS}~\cite{dai2024siggraph-gaussian-surfels}: Gaussian surfels splatting, a state-of-the-art 2D Gaussian representation using surfels instead of 3D ellipsoids. It uses the same setup as the \textit{3DGS} baseline but adds the depth-normal consistency loss from GSS~\cite{dai2024siggraph-gaussian-surfels}.

\item \textit{Neural Point+3DGS}: Our extension of Scaffold-GS~\cite{lu2024cvpr-scaffoldgs} that enables incremental training and adds supervision of neural point geometric features through the SDF branch, as detailed in \secref{subsec:ngsf}.

\item \textit{Neural Point+GSS}: A variant that replaces the 3D Gaussian in \textit{Neural Point+3DGS} with 2D Gaussian surfels.

\item \textit{PINGS}: Our complete framework that extends \textit{Neural Point+GSS} by introducing geometric consistency loss $\mathcal{L}_{\text{cons}}$ into the joint training, as described in \secref{subsec:cons}.

\end{itemize}

For fair comparison, we disable the localization part and use the ground truth pose for all the compared methods. 
For 3DGS and GSS, we initialize their Gaussian primitive density to match the total number of Gaussians spawned by PINGS.

We show the quantitative comparison on five sequences in \tabref{tab:rendering} as well as show the qualitative comparison of the RGB and normal map rendering results on the \textit{church} scene at a novel view in \figref{fig:qualitative_rendering_church_1}. 
Our method PINGS achieves superior performance in both photorealistic rendering quality and depth rendering  accuracy on the in-house car dataset, and consistently outperforms the baselines for both in-sequence and out-of-sequence testing views.
Analysis of the results reveals several insights:
(i) The adoption of GSS over 3DGS leads to improved geometric rendering quality and enhanced out-of-sequence rendering photorealism;
(ii) Our approach of spawning Gaussians from neural points and jointly training with the distance field provides better optimization control and reduces floating Gaussians in free space, resulting in superior rendering quality;
(iii) The addition of geometric consistency constraints from SDF enables better surface alignment of Gaussian surfels, further enhancing both geometric accuracy and photorealistic rendering quality, as evidenced by the smoother normal maps produced by PINGS compared to \textit{Neural Point+GSS}.
These improvements are less significant in the \textit{Roundabout} scene, where the dense viewpoint coverage from the vehicle's circular trajectory provides strong multi-view constraints, reducing the benefit of additional geometric constraints from the SDF.

In sum, this experiment validates that PINGS achieves better RGB and geometric rendering at novel views by constraining the Gaussian splatting radiance field using the SDF.

\subsection{Surface Reconstruction Quality Evaluation}


We evaluate surface reconstruction quality on four sequences from the Oxford-Spires dataset.
We follow the benchmark~\cite{tao2024arxiv-oxford} to report the metrics including accuracy error, completeness error, and Chamfer distance, as well as precision, recall and F-score calculated with a threshold of \SI{0.1}{\meter}.
We compare the performance of PINGS with five state-of-the-art methods, including OpenMVS~\cite{cernea2020openmvs}, Nerfacto~\cite{tancik2023siggraph-nerfstudio}, GSS~\cite{dai2024siggraph-gaussian-surfels}, VDB-Fusion~\cite{vizzo2022sensors}, and PIN-SLAM~\cite{pan2024tro}.
%
To ensure fair comparison of geometric mapping quality, we disable the localization modules of PIN-SLAM and PINGS and use ground truth poses across all methods.
For GSS, after completing the radiance field mapping, we render depth maps at each frame and apply TSDF fusion~\cite{vizzo2022sensors} for mesh extraction.
%
%
Results of the vision-based offline processing methods (OpenMVS and Nerfacto) are taken from the benchmark~\cite{tao2024arxiv-oxford}.
For the remaining methods (GSS, VDB-Fusion, PIN-SLAM, and PINGS), we extract surface meshes from their SDFs using marching cubes~\cite{lorensen1987siggraph} at a resolution of \SI{0.1}{\meter}.

We show the qualitative results of PINGS on the four sequences in \figref{fig:qualitative_surface_reconstruction}.
Quantitative comparisons in \tabref{tab:recon_table} demonstrate that PINGS achieves superior performance, particularly in terms of Chamfer distance and F-score metrics.
Notably, when using identical neural point resolution, PINGS consistently outperforms PIN-SLAM across all metrics through its joint optimization of the radiance field and geometric consistency constraints.
This improvement validates that incorporating dense photometric cues and multi-view consistency from the radiance field improves the SDF accuracy, ultimately enabling surface mesh reconstruction with higher quality.

\subsection{SLAM Localization Accuracy Evaluation}

We compare the pose estimation performance of PINGS against state-of-the-art LiDAR odometry/SLAM systems on two full sequences of the in-house car dataset. 
The compared methods include F-LOAM~\cite{wang2021iros-fflo}, KISS-ICP~\cite{vizzo2023ral}, SuMa~\cite{behley2018rss}, MULLS~\cite{pan2021icra-mvls}, and PIN-SLAM~\cite{pan2024tro}.
For evaluation metrics, we use average relative translation error (ARTE)~\cite{geiger2012cvpr} to assess odometry drift and absolute trajectory error (ATE)~\cite{zhang2018iros-evo} to measure the global pose estimation accuracy.
The results shown in \tabref{tab:localization_comparison} demonstrate that PINGS achieves both lower odometry drift and superior global localization accuracy than the compared approaches.
Compared to PIN-SLAM, the improvement stems from the refined SDF obtained through joint optimization with the radiance field and geometric consistency constraints.
The improved SDF leads to more accurate LiDAR odometry and relocalization during loop closure correction.

\begin{table}[!t]
\centering
\caption{Localization performance comparison of PINGS against state-of-the-art odometry/SLAM methods on the \textit{in-house car dataset}. We report average relative translation error (ARTE) [\%] and absolute trajectory error (ATE) [m]. Odometry methods are shown above the midrule, SLAM methods below. Best results are shown in \textbf{bold}, second best are \underline{underscored}.}
\label{tab:localization_comparison}
\resizebox{0.48\textwidth}{!}{
\begin{tabular}{l|cc|cc}
  \toprule
    \multirow{2}{*}{Method}  & \multicolumn{2}{c|}{\text{Seq. 1 (\SI{5.0}{\km})}} & \multicolumn{2}{c}{\text{Seq. 2 (\SI{3.7}{\km})}} \\
     & ARTE [\%] $\downarrow$ & ATE [m] $\downarrow$ & ARTE [\%] $\downarrow$ & ATE [m] $\downarrow$ \\
    \midrule
    F-LOAM~\cite{wang2021iros-fflo} & 1.96 & 28.52 & 1.93 & 27.00 \\
    KISS-ICP~\cite{vizzo2023ral} & 1.49 & 8.17 & 1.38 & 8.22 \\
    PIN odometry~\cite{pan2024tro} & 0.95 & 4.51 & 0.98 & 5.64 \\
    PINGS odometry & \underline{0.73} & 5.17 & \underline{0.59} & 4.78  \\
    \midrule
    SuMa~\cite{behley2018rss} & 5.55 & 39.90 & 4.42 & 44.78 \\
    MULLS~\cite{pan2021icra-mvls} & 2.23 & 40.37 & 1.64 & 33.82 \\
    PIN-SLAM~\cite{pan2024tro} & 1.00 & \underline{3.17} & 0.98 & \underline{4.44} \\
    PINGS (Ours) & \bf{0.68} & \bf{1.99} & \bf{0.58} & \bf{3.47} \\
    \bottomrule
\end{tabular}
}
\vspace{-10pt}
\end{table}

\subsection{Large-Scale Globally Consistent Mapping}

\begin{figure}[!t]
  \centering  
  \includegraphics[width=0.99\linewidth]{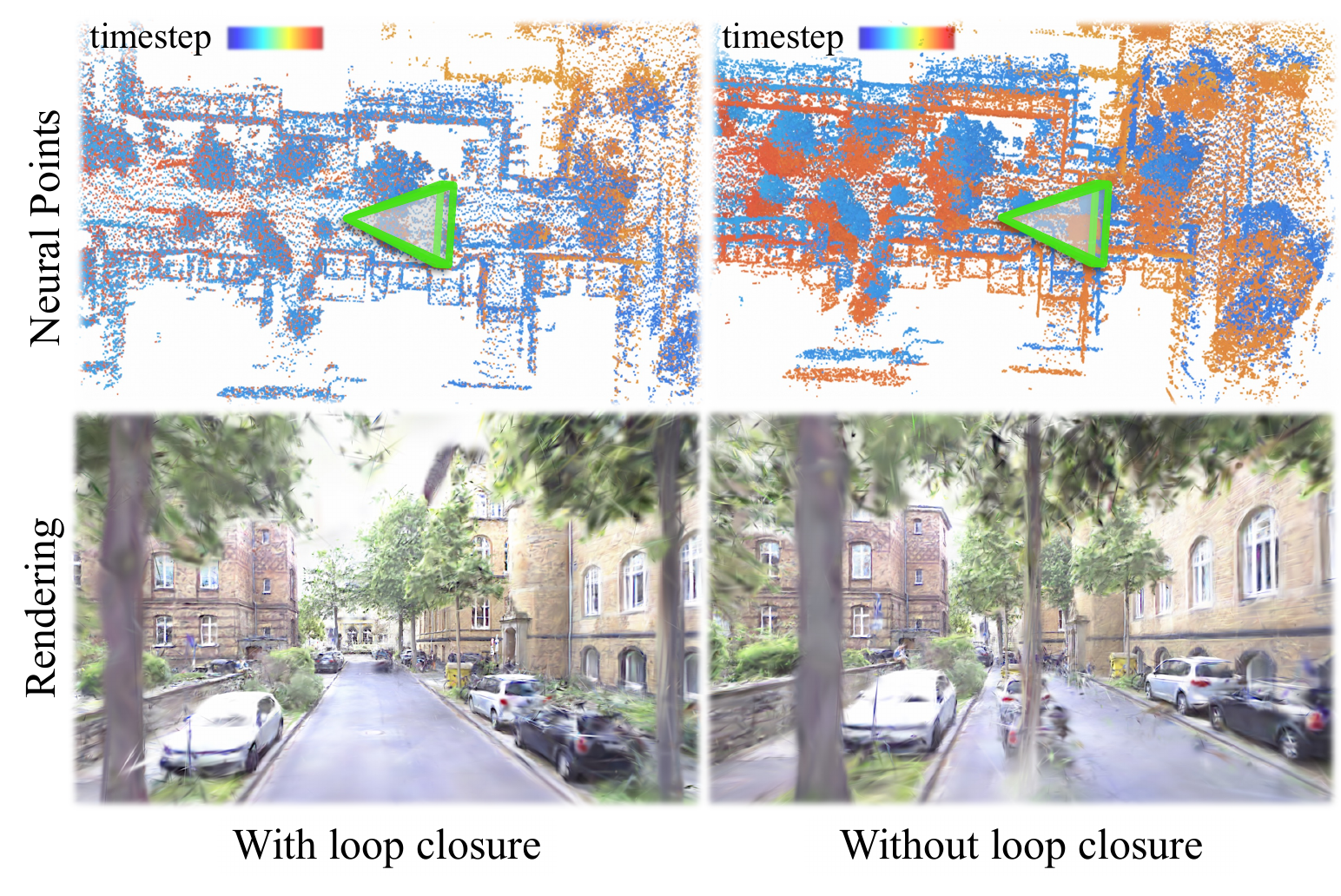}
  \setlength{\abovecaptionskip}{-12pt}
  \caption{
    Demonstration of the effect of loop closure correction on the \textit{in-house car dataset}. 
    When the vehicle revisits a previously mapped region, we compare the neural point map (colored by timestep) and RGB rendering (viewed from the green frustum) with and without correcting the loop closure. 
    Without loop closure correction, the misaligned neural points create visual artifacts like trees appearing on the road. 
    After applying loop closure correction and map update, we achieve globally consistent neural point map and RGB rendering.  
    }
  \label{fig:loop_closure}
  \vspace{-8pt}
\end{figure}

\figref{fig:teaser} demonstrates the globally consistent SLAM capabilities of PINGS on a challenging \SI{5}{\km} sequence from our in-house car dataset.
In \figref{fig:loop_closure}, we show the effect of loop closure correction. 
Without loop closure correction, odometry drift accumulates over time, causing neural points to be inconsistently placed when revisiting previously mapped regions. 
This results in visual artifacts in the radiance field rendering, such as duplicate objects and trees incorrectly appearing on the road.
After conducting loop closure correction and updating the map, both the neural point map and RGB rendering achieve global consistency.
These results validate that PINGS can build globally consistent maps at large scale through loop closure correction by leveraging the elasticity of neural points.

\begin{figure}[tbp]
  \centering  
  \subfigure{\includegraphics[height=0.16\textwidth]{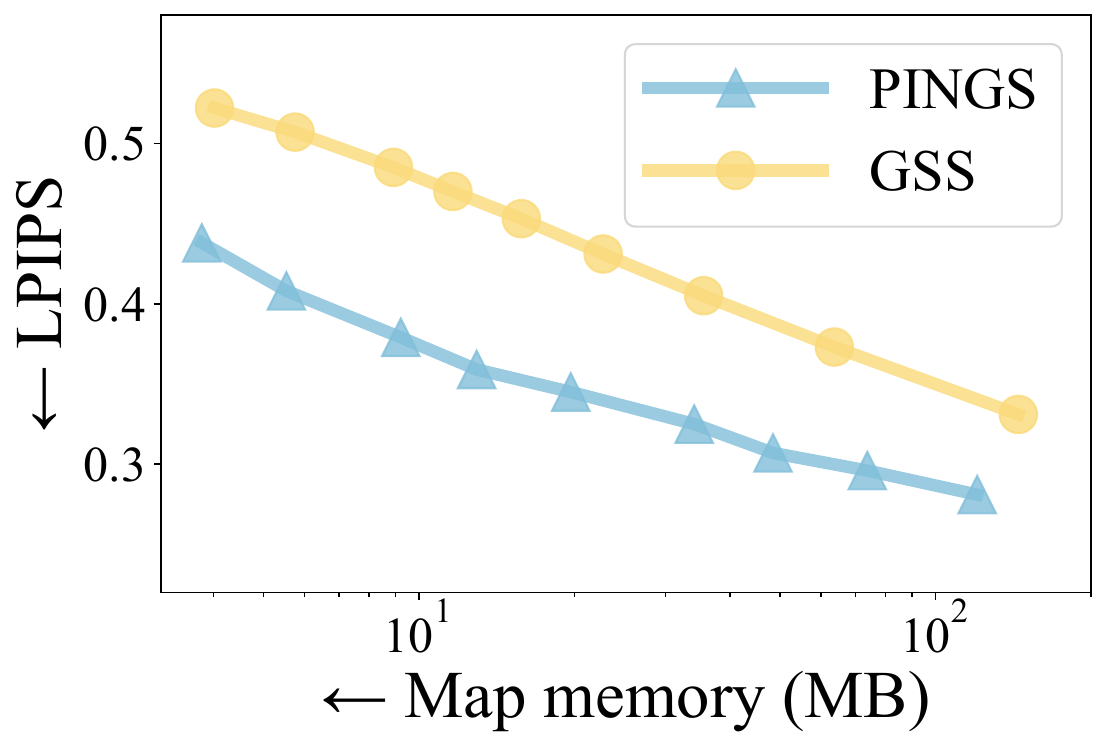}}
  \subfigure{\includegraphics[height=0.16\textwidth]{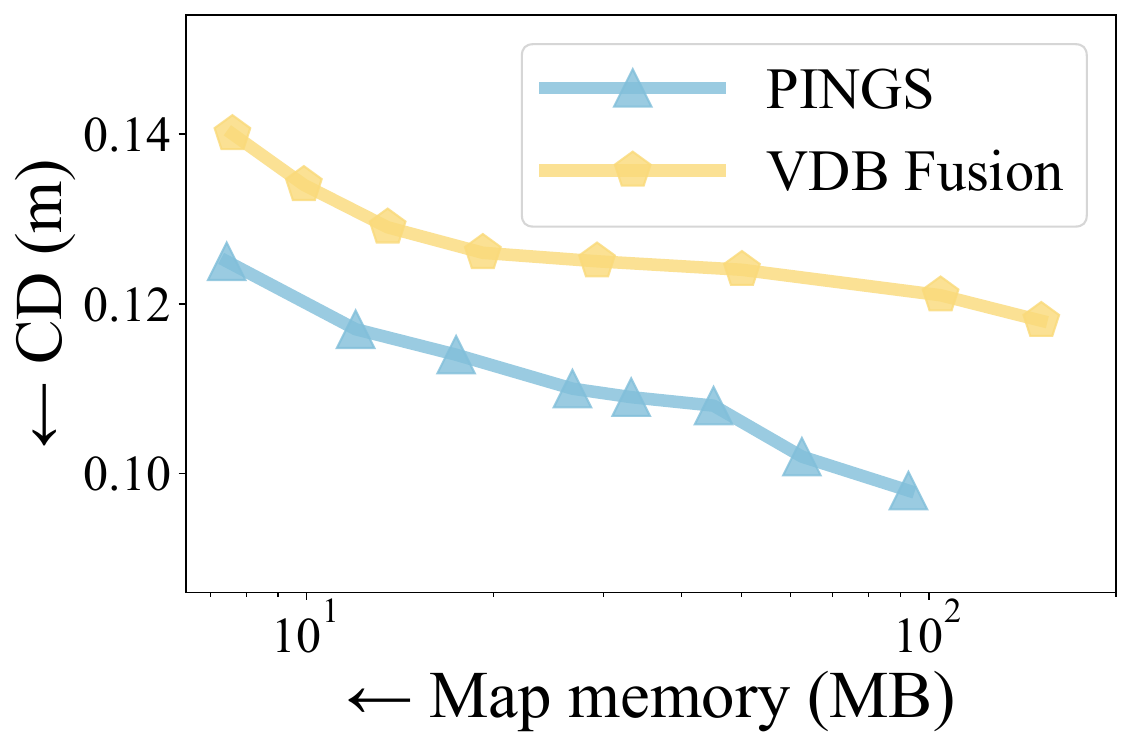}}
  \setlength{\abovecaptionskip}{-8pt}
  \caption{Map memory efficiency analysis comparing mapping quality versus memory usage. Left: Radiance field comparison between PINGS and GSS on the \textit{in-house car dataset} using LPIPS metric. Right: Distance field comparison between PINGS and VDB-Fusion on the \textit{Oxford Spires dataset} using Chamfer distance. Points represent results at different map resolution, with points closer to the bottom-left corner indicating better quality-memory trade-off.}
  \label{fig:map_memory}
  \vspace{-8pt}
\end{figure}

\subsection{Map Memory Efficiency Evaluation}

\figref{fig:map_memory} depicts the map memory usage in relation to the rendering quality for the radiance field and the surface reconstruction quality for the distance field. 
%
%
Experiment results validate that storing the neural points and decoder MLPs in PINGS is more memory-efficient than directly storing the Gaussian primitives or the discrete SDF in voxels.
With equivalent memory usage, PINGS achieves superior performance across both metrics: better novel view rendering photorealism (lower LPIPS) compared to GSS~\cite{dai2024siggraph-gaussian-surfels} on the in-house car dataset, and better surface reconstruction accuracy (lower Chamfer distance) compared to the discrete TSDF-based method VDB-Fusion~\cite{vizzo2022sensors} on the Oxford Spires dataset.
Moreover, while GSS and VDB-Fusion each model only a single field type, our PINGS framework efficiently represents both radiance field and SDF within a single map.
%
The efficiency of PINGS comes from the globally-shared decoder MLPs that learn common patterns, and locally-defined neural points that compactly encode multiple Gaussian primitives and continuous SDF values through feature vectors instead of storing them explicitly.

\section{Limitations}
\label{sec:limitations}

Our current approach has three main limitations. 
First, although our SDF mapping and LiDAR odometry modules operate at sensor frame rate, the computationally intensive radiance field mapping results in an overall processing time of around five seconds per frame on an NVIDIA A6000 GPU.
This performance bottleneck could potentially be addressed through recent advances in efficient optimization schemes for Gaussian splatting training~\cite{hoellein2024arxiv-3dgslm, mallick2024siggraph-taming3dgs}.
Second, PINGS relies solely on online per-scene optimization without using any pre-trained priors. 
Incorporating such priors~\cite{chen2024eccv-g3r} into our unified map representation could improve both mapping quality and convergence speed.
%
Finally, though PINGS can filter dynamic objects using the distance field, it lacks explicit 4D modeling capabilities. 
This limitation is noticeable in highly dynamic environments and when objects transition between static and dynamic states. 
Future work could address this challenge by incorporating object detection priors~\cite{yan2024eccv-streetgs, chen2024arxiv-omnire} to enable accurate 4D mapping of both radiance and distance fields.
%

\section{Conclusion}
\label{sec:conclusion}

In this paper, we present a new LiDAR-visual SLAM system making use of a novel map representation that unifies a continuous signed distance field and a Gaussian splatting radiance field within an elastic and compact set of neural points.
By introducing mutual geometric consistency constraints between these fields, we jointly improve both representations.
The distance field provides geometric structure to guide radiance field optimization, while the radiance field's dense photometric cues and multi-view consistency enhance the distance field's accuracy.
Our experimental results on challenging large-scale datasets show that our method can incrementally construct globally consistent maps that outperform baseline methods in the novel view rendering fidelity, surface reconstruction quality, odometry estimation accuracy, and map memory efficiency.


\section*{Acknowledgements}
\label{sec:acknowledgements}
This work has partially been funded by the European Union under the grant agreements No~101070405~(DigiForest), by the Deutsche Forschungsgemeinschaft (DFG, German Research Foundation) under Germany's Excellence Strategy, EXC-2070 -- 390732324 -- PhenoRob, under STA~1051/5-1 within the FOR 5351~--~459376902~(AID4Crops), and by the German Federal Ministry
of Education and Research (BMBF) in the project ``Robotics Institute Germany", grant No. 16ME0999.

\vspace{8pt}


\bibliographystyle{plain_abbrv}
\bibliography{glorified,new}


\end{document}

%% file: stachnisslab-latex.tex
\usepackage{graphics}           
\usepackage{times}              
\usepackage{amsmath}            
\usepackage{amssymb}            
\usepackage{graphicx}
\usepackage{algorithm}
\usepackage[noend]{algpseudocode}
\usepackage{booktabs}
\usepackage{color}
\usepackage[nocompress]{cite} 
\definecolor{instructioncolor}{rgb}{.5,.5,.5}

\usepackage[font=small]{caption}

\def\secref#1{Sec.~\ref{#1}}
\def\figref#1{Fig.~\ref{#1}}
\def\tabref#1{Tab.~\ref{#1}}
\def\eqref#1{Eq.~(\ref{#1})}

\makeatletter
\usepackage{xspace}
\DeclareRobustCommand\onedot{\futurelet\@let@token\@onedot}
\def\@onedot{\ifx\@let@token.\else.\null\fi\xspace}

\def\etal{{et al}\onedot}
\makeatother

\def\etalcite#1{\etal~\cite{#1}}

\usepackage{array}
\newcolumntype{L}[1]{>{\raggedright\let\newline\\\arraybackslash\hspace{0pt}}m{#1}}
\newcolumntype{C}[1]{>{\centering\let\newline\\\arraybackslash\hspace{0pt}}m{#1}}
\newcolumntype{R}[1]{>{\raggedleft\let\newline\\\arraybackslash\hspace{0pt}}m{#1}}

%% file: stachnisslab-math.tex
\newcommand{\RR}{\mathbb{R}}

\newcommand{\norm}[1]{\lVert#1\lVert}

\newcommand{\mybold}[1]{\mbox{\boldmath$#1$}}

\newcommand{\vv}[1]{{\mybold #1}} 

\newcommand{\m}[1]{{\mbox{{\sffamily\slshape{#1\/}}}}}

\newcommand{\mq}[1]{{\mbox{{\sffamily{#1}}}}}

\newcommand{\tr}[0]{\sf T}              










%% file: pan2025rss_arxiv_v2.bbl
\begin{thebibliography}{10}

\bibitem{abou-chakra2024wacv}
J.~Abou-Chakra, F.~Dayoub, and N.~S\"underhauf.
\newblock {ParticleNeRF: A Particle-Based Encoding for Online Neural Radiance
  Fields}.
\newblock In {\em Proc.~of the IEEE Winter Conf.~on Applications of Computer
  Vision (WACV)}, 2024.

\bibitem{abou-chakra2024corl}
J.~Abou-Chakra, K.~Rana, F.~Dayoub, and N.~Suenderhauf.
\newblock {Physically Embodied Gaussian Splatting: A Realtime Correctable World
  Model for Robotics}.
\newblock In {\em Proc.~of the Conf.~on Robot Learning (CoRL)}, 2024.

\bibitem{behley2018rss}
J.~Behley and C.~Stachniss.
\newblock {Efficient Surfel-Based SLAM using 3D Laser Range Data in Urban
  Environments}.
\newblock In {\em Proc.~of Robotics: Science and Systems (RSS)}, 2018.

\bibitem{bortolon2024eccv-6dgs}
M.~Bortolon, T.~Tsesmelis, S.~James, F.~Poiesi, and A.~{Del Bue}.
\newblock {6DGS: 6D Pose Estimation from a Single Image and a 3D Gaussian
  Splatting Model}.
\newblock In {\em Proc.~of the Europ.~Conf.~on Computer Vision (ECCV)}, 2024.

\bibitem{cernea2020openmvs}
D.~Cernea.
\newblock {OpenMVS: Multi-View Stereo Reconstruction Library}, 2020.

\bibitem{chen2023arxiv-neusg}
H.~Chen, C.~Li, and G.H. Lee.
\newblock {NeuSG: Neural implicit surface reconstruction with 3d gaussian
  splatting guidance}.
\newblock {\em arXiv preprint}, arXiv:2023.00846, 2023.

\bibitem{chen2024eccv-g3r}
Y.~Chen, J.~Wang, Z.~Yang, S.~Manivasagam, and R.~Urtasun.
\newblock {G3R: Gradient Guided Generalizable Reconstruction}.
\newblock In {\em Proc.~of the Europ.~Conf.~on Computer Vision (ECCV)}, 2024.

\bibitem{chen2023iccv-neurbf}
Z.~Chen, Z.~Li, L.~Song, L.~Chen, J.~Yu, J.~Yuan, and Y.~Xu.
\newblock {NeuRBF: A Neural Fields Representation with Adaptive Radial Basis
  Functions}.
\newblock In {\em Proc.~of the IEEE/CVF Intl.~Conf.~on Computer Vision (ICCV)},
  2023.

\bibitem{chen2024arxiv-omnire}
Z.~Chen, J.~Yang, J.~Huang, R.d. Lutio, J.M. Esturo, B.~Ivanovic, O.~Litany,
  Z.~Gojcic, S.~Fidler, M.~Pavone, L.~Song, and Y.~Wang.
\newblock {OmniRe: Omni Urban Scene Reconstruction}.
\newblock {\em arXiv preprint}, arXiv:2408.16760, 2024.

\bibitem{cui2024tog-letsgo}
J.~Cui, J.~Cao, F.~Zhao, Z.~He, Y.~Chen, Y.~Zhong, L.~Xu, Y.~Shi, and J.~Yu.
\newblock {LetsGo: Large-Scale Garage Modeling and Rendering via LiDAR Assisted
  Gaussian Primitives}.
\newblock {\em ACM Trans.~on Graphics (TOG)}, 43(6):1--18, 2024.

\bibitem{dai2024siggraph-gaussian-surfels}
P.~Dai, J.~Xu, W.~Xie, X.~Liu, H.~Wang, and W.~Xu.
\newblock {High-quality Surface Reconstruction using Gaussian Surfels}.
\newblock In {\em Proc.~of the Intl.~Conf.~on Computer Graphics and Interactive
  Techniques (SIGGRAPH)}, 2024.

\bibitem{driess2022neurips-nerfrl}
D.~Driess, I.~Schubert, P.~Florence, Y.~Li, and M.~Toussaint.
\newblock {Reinforcement Learning with Neural Radiance Fields}.
\newblock In {\em Proc.~of the Conf. on Neural Information Processing Systems
  (NeurIPS)}, 2022.

\bibitem{fischer2024neurips-dgf}
T.~Fischer, J.~Kulhanek, S.R. Bul{\`o}, L.~Porzi, M.~Pollefeys, and
  P.~Kontschieder.
\newblock {Dynamic 3D Gaussian Fields for Urban Areas}.
\newblock In {\em Proc.~of the Conf. on Neural Information Processing Systems
  (NeurIPS)}, 2024.

\bibitem{fox1997jra}
D.~Fox, W.~Burgard, and S.~Thrun.
\newblock The dynamic window approach to collision avoidance.
\newblock {\em IEEE Journal of Robotics and Automation}, 4(1):23--33, 1997.

\bibitem{geiger2012cvpr}
A.~Geiger, P.~Lenz, and R.~Urtasun.
\newblock {Are we ready for Autonomous Driving? The KITTI Vision Benchmark
  Suite}.
\newblock In {\em Proc.~of the IEEE Conf.~on Computer Vision and Pattern
  Recognition (CVPR)}, 2012.

\bibitem{grisetti2007tro}
G.~Grisetti, C.~Stachniss, and W.~Burgard.
\newblock {Improved Techniques for Grid Mapping with Rao-Blackwellized Particle
  Filters}.
\newblock {\em IEEE Trans.~on Robotics (TRO)}, 23(1):34--46, 2007.

\bibitem{gropp2020icml}
A.~Gropp, L.~Yariv, N.~Haim, M.~Atzmon, and Y.~Lipman.
\newblock {Implicit Geometric Regularization for Learning Shapes}.
\newblock In {\em Proc.~of the Intl.~Conf.~on Machine Learning (ICML)}, 2020.

\bibitem{guedon2024cvpr-sugar}
A.~Gu{\'e}don and V.~Lepetit.
\newblock {Sugar: Surface-aligned gaussian splatting for efficient 3d mesh
  reconstruction and high-quality mesh rendering}.
\newblock In {\em Proc.~of the IEEE/CVF Conf.~on Computer Vision and Pattern
  Recognition (CVPR)}, 2024.

\bibitem{hess2024arxiv-splatad}
G.~Hess, C.~Lindstr{\"o}m, M.~Fatemi, C.~Petersson, and L.~Svensson.
\newblock {SplatAD: Real-Time Lidar and Camera Rendering with 3D Gaussian
  Splatting for Autonomous Driving}.
\newblock {\em arXiv preprint}, arXiv:2411.16816, 2024.

\bibitem{hoellein2024arxiv-3dgslm}
L.~H{\"o}llein, A.~Bo\v{z}i\v{c}, M.~Zollh{\"o}fer, and M.~Nie{\ss}ner.
\newblock {3DGS-LM: Faster Gaussian-Splatting Optimization with
  Levenberg-Marquardt}.
\newblock {\em arXiv preprint}, arXiv:2409.12892, 2024.

\bibitem{hong2024ral-livgaussmap}
S.~Hong, J.~He, X.~Zheng, C.~Zheng, and S.~Shen.
\newblock {LIV-Gaussmap: Lidar-inertial-visual fusion for real-time 3d radiance
  field map rendering}.
\newblock {\em IEEE Robotics and Automation Letters (RA-L)}, 9(11):9765--9772,
  2024.

\bibitem{hornung2013ar}
A.~Hornung, K.~Wurm, M.~Bennewitz, C.~Stachniss, and W.~Burgard.
\newblock {OctoMap: An Efficient Probabilistic 3D Mapping Framework Based on
  Octrees}.
\newblock {\em Autonomous Robots}, 34(3):189--206, 2013.

\bibitem{huang2024siggraph-2dgs}
B.~Huang, Z.~Yu, A.~Chen, A.~Geiger, and S.~Gao.
\newblock {2D Gaussian Splatting for Geometrically Accurate Radiance Fields}.
\newblock In {\em Proc.~of the Intl.~Conf.~on Computer Graphics and Interactive
  Techniques (SIGGRAPH)}, 2024.

\bibitem{hughes2024ijrr}
N.~Hughes, Y.~Chang, S.~Hu, R.~Talak, R.~Abdulhai, J.~Strader, and L.~Carlone.
\newblock {Foundations of spatial perception for robotics: Hierarchical
  representations and real-time systems}.
\newblock {\em Intl.~Journal~of Robotics Research (IJRR)}, 43(10):1457--1505,
  2024.

\bibitem{jiang2024arxiv-ligs}
C.~Jiang, R.~Gao, K.~Shao, Y.~Wang, R.~Xiong, and Y.~Zhang.
\newblock {LI-GS: Gaussian Splatting with LiDAR Incorporated for Accurate
  Large-Scale Reconstruction}.
\newblock {\em arXiv preprint}, arXiv:2409.12899, 2024.

\bibitem{jin2025ral-activegs}
L.~Jin, X.~Zhong, Y.~Pan, J.~Behley, C.~Stachniss, and M.~Popovic.
\newblock {ActiveGS: Active Scene Reconstruction using Gaussian Splatting}.
\newblock {\em IEEE Robotics and Automation Letters (RA-L)}, 10(5):4866--4873,
  2025.

\bibitem{jin2024iros-gsplanner}
R.~Jin, Y.~Gao, H.~Lu, and F.~Gao.
\newblock {GS-Planner: A Gaussian-Splatting-based Planning Framework for Active
  High-Fidelity Reconstruction}.
\newblock In {\em Proc.~of the IEEE/RSJ Intl.~Conf.~on Intelligent Robots and
  Systems (IROS)}, 2024.

\bibitem{kazhdan2013acmgraphics}
M.~Kazhdan and H.~Hoppe.
\newblock Screened poisson surface reconstruction.
\newblock {\em ACM Trans.~on Graphics}, 32(3):1--13, 2013.

\bibitem{keetha2024cvpr-splatam}
N.~Keetha, J.~Karhade, K.M. Jatavallabhula, G.~Yang, S.~Scherer, D.~Ramanan,
  and J.~Luiten.
\newblock {SplaTAM: Splat Track \& Map 3D Gaussians for Dense RGB-D SLAM}.
\newblock In {\em Proc.~of the IEEE/CVF Conf.~on Computer Vision and Pattern
  Recognition (CVPR)}, 2024.

\bibitem{kerbl2023tog-3dgs}
B.~Kerbl, G.~Kopanas, T.~Leimk{\"u}hler, and G.~Drettakis.
\newblock {3D Gaussian Splatting for Real-Time Radiance Field Rendering}.
\newblock {\em ACM Trans.~on Graphics (TOG)}, 42(4):1--14, 2023.

\bibitem{kerbl2024tog-hierarchical3dgs}
B.~Kerbl, A.~Meuleman, G.~Kopanas, M.~Wimmer, A.~Lanvin, and G.~Drettakis.
\newblock {A Hierarchical 3D Gaussian Representation for Real-Time Rendering of
  Very Large Datasets}.
\newblock {\em ACM Trans.~on Graphics (TOG)}, 43(4):1--15, 2024.

\bibitem{kingma2015iclr}
D.~Kingma and J.~Ba.
\newblock {Adam: {A} Method for Stochastic Optimization}.
\newblock In {\em Proc.~of the Intl.~Conf.~on Learning Representations (ICLR)},
  2015.

\bibitem{klingensmith2015rss}
M.~Klingensmith, I.~Dryanovski, S.~Srinivasa, and J.~Xiao.
\newblock Chisel: Real time large scale 3d reconstruction onboard a mobile
  device using spatially hashed signed distance fields.
\newblock In {\em Proc.~of Robotics: Science and Systems (RSS)}, 2015.

\bibitem{li2022cvpr-dccdif}
T.~Li, X.~Wen, Y.S. Liu, H.~Su, and Z.~Han.
\newblock {Learning Deep Implicit Functions for 3D Shapes with Dynamic Code
  Clouds}.
\newblock In {\em Proc.~of the IEEE/CVF Conf.~on Computer Vision and Pattern
  Recognition (CVPR)}, 2022.

\bibitem{li2024arxiv-activesplat}
Y.~Li, Z.~Kuang, T.~Li, G.~Zhou, S.~Zhang, and Z.~Yan.
\newblock {ActiveSplat: High-Fidelity Scene Reconstruction through Active
  Gaussian Splatting}.
\newblock {\em arXiv preprint}, arXiv:2410.21955, 2024.

\bibitem{liu2024eccv-citygaussian}
Y.~Liu, C.~Luo, L.~Fan, N.~Wang, J.~Peng, and Z.~Zhang.
\newblock {Citygaussian: Real-time high-quality large-scale scene rendering
  with gaussians}.
\newblock In {\em Proc.~of the Europ.~Conf.~on Computer Vision (ECCV)}, 2024.

\bibitem{lorensen1987siggraph}
W.~Lorensen and H.~Cline.
\newblock {Marching Cubes: a High Resolution 3D Surface Construction
  Algorithm}.
\newblock In {\em Proc.~of the Intl.~Conf.~on Computer Graphics and Interactive
  Techniques (SIGGRAPH)}, 1987.

\bibitem{lu2024cvpr-scaffoldgs}
T.~Lu, M.~Yu, L.~Xu, Y.~Xiangli, L.~Wang, D.~Lin, and B.~Dai.
\newblock {Scaffold-GS: Structured 3d gaussians for view-adaptive rendering}.
\newblock In {\em Proc.~of the IEEE/CVF Conf.~on Computer Vision and Pattern
  Recognition (CVPR)}, 2024.

\bibitem{lyu2024tog-3dgsr}
X.~Lyu, Y.T. Sun, Y.H. Huang, X.~Wu, Z.~Yang, Y.~Chen, J.~Pang, and X.~Qi.
\newblock {3DGSR: Implicit Surface Reconstruction with 3D Gaussian Splatting}.
\newblock {\em ACM Trans.~on Graphics (TOG)}, 43(6):1--12, 2024.

\bibitem{mallick2024siggraph-taming3dgs}
S.S. Mallick, R.~Goel, B.~Kerbl, M.~Steinberger, F.V. Carrasco, and
  F.~De~La~Torre.
\newblock {Taming 3DGS: High-Quality Radiance Fields with Limited Resources}.
\newblock In {\em Proc.~of the Intl.~Conf.~on Computer Graphics and Interactive
  Techniques (SIGGRAPH)}, 2024.

\bibitem{mascaro2024annurevcontrol}
R.~Mascaro and M.~Chli.
\newblock {Scene Representations for Robotic Spatial Perception}.
\newblock {\em Annual Review of Control, Robotics, and Autonomous Systems},
  8(5):1--27, 2024.

\bibitem{matsuki2024cvpr-monogs}
H.~Matsuki, R.~Murai, P.H. Kelly, and A.J. Davison.
\newblock {Gaussian splatting SLAM}.
\newblock In {\em Proc.~of the IEEE/CVF Conf.~on Computer Vision and Pattern
  Recognition (CVPR)}, 2024.

\bibitem{wildersmith2024iros}
M.H. Maximum Wilder-Smith, Vaishakh~Patil.
\newblock {Radiance Fields for Robotic Teleoperation}.
\newblock In {\em Proc.~of the IEEE/RSJ Intl.~Conf.~on Intelligent Robots and
  Systems (IROS)}, 2024.

\bibitem{mescheder2019cvpr}
L.~Mescheder, M.~Oechsle, M.~Niemeyer, S.~Nowozin, and A.~Geiger.
\newblock {Occupancy networks: Learning 3d reconstruction in function space}.
\newblock In {\em Proc.~of the IEEE/CVF Conf.~on Computer Vision and Pattern
  Recognition (CVPR)}, 2019.

\bibitem{mildenhall2020eccv}
B.~Mildenhall, P.~Srinivasan, M.~Tancik, J.~Barron, R.~Ramamoorthi, and R.~Ng.
\newblock {NeRF: Representing Scenes as Neural Radiance Fields for View
  Synthesis}.
\newblock In {\em Proc.~of the Europ.~Conf.~on Computer Vision (ECCV)}, 2020.

\bibitem{newcombe2011ismar}
R.A. Newcombe, S.~Izadi, O.~Hilliges, D.~Molyneaux, D.~Kim, A.J. Davison,
  P.~Kohli, J.~Shotton, S.~Hodges, and A.~Fitzgibbon.
\newblock {KinectFusion: Real-Time Dense Surface Mapping and Tracking}.
\newblock In {\em Proc.~of the Intl.~Symp.~on Mixed and Augmented Reality
  (ISMAR)}, 2011.

\bibitem{niessner2013tog}
M.~Nie\ss{}ner, M.~Zollh\"{o}fer, S.~Izadi, and M.~Stamminger.
\newblock {Real-Time 3D Reconstruction at Scale Using Voxel Hashing}.
\newblock {\em ACM Trans.~on Graphics (TOG)}, 32(6):1--11, 2013.

\bibitem{oleynikova2017iros}
H.~Oleynikova, Z.~Taylor, M.~Fehr, R.~Siegwar, and J.~Nieto.
\newblock Voxblox: Incremental 3d euclidean signed distance fields for on-board
  mav planning.
\newblock In {\em Proc.~of the IEEE/RSJ Intl.~Conf.~on Intelligent Robots and
  Systems (IROS)}, 2017.

\bibitem{ortiz2022rss}
J.~Ortiz, A.~Clegg, J.~Dong, E.~Sucar, D.~Novotny, M.~Zollhoefer, and
  M.~Mukadam.
\newblock isdf: Real-time neural signed distance fields for robot perception.
\newblock In {\em Proc.~of Robotics: Science and Systems (RSS)}, 2022.

\bibitem{pan2021icra-mvls}
Y.~Pan, P.~Xiao, Y.~He, Z.~Shao, and Z.~Li.
\newblock {MULLS: Versatile LiDAR SLAM Via Multi-Metric Linear Least Square}.
\newblock In {\em Proc.~of the IEEE Intl.~Conf.~on Robotics \& Automation
  (ICRA)}, 2021.

\bibitem{pan2024tro}
Y.~Pan, X.~Zhong, L.~Wiesmann, T.~Posewsky, J.~Behley, and C.~Stachniss.
\newblock {PIN-SLAM: LiDAR SLAM Using a Point-Based Implicit Neural
  Representation for Achieving Global Map Consistency}.
\newblock {\em IEEE Trans.~on Robotics (TRO)}, 40:4045--4064, 2024.

\bibitem{park2019cvpr}
J.J. Park, P.~Florence, J.~Straub, R.~Newcombe, and S.~Lovegrove.
\newblock {DeepSDF: Learning Continuous Signed Distance Functions for Shape
  Representation}.
\newblock In {\em Proc.~of the IEEE/CVF Conf.~on Computer Vision and Pattern
  Recognition (CVPR)}, 2019.

\bibitem{reijgwart2023rss-wavemap}
V.~Reijgwart, C.~Cadena, R.~Siegwart, and L.~Ott.
\newblock {Efficient volumetric mapping of multi-scale environments using
  wavelet-based compression}.
\newblock In {\em Proc.~of Robotics: Science and Systems (RSS)}, 2023.

\bibitem{rematas2022cvpr}
K.~Rematas, A.~Liu, P.P. Srinivasan, J.T. Barron, A.~Tagliasacchi,
  T.~Funkhouser, and V.~Ferrari.
\newblock Urban radiance fields.
\newblock In {\em Proc.~of the IEEE/CVF Conf.~on Computer Vision and Pattern
  Recognition (CVPR)}, 2022.

\bibitem{ren2024arxiv-octreegs}
K.~Ren, L.~Jiang, T.~Lu, M.~Yu, L.~Xu, Z.~Ni, and B.~Dai.
\newblock {Octree-GS: Towards Consistent Real-time Rendering with
  LOD-structured 3D Gaussians}.
\newblock {\em arXiv preprint}, arXiv:2403.17898, 2024.

\bibitem{sandstrom2023iccv-pointslam}
E.~Sandström, Y.~Li, L.~Van~Gool, and M.~R.~Oswald.
\newblock {Point-SLAM: Dense Neural Point Cloud-based SLAM}.
\newblock In {\em Proc.~of the IEEE/CVF Intl.~Conf.~on Computer Vision (ICCV)},
  2023.

\bibitem{schmid2023ral-dynablox}
L.~Schmid, O.~Andersson, A.~Sulser, P.~Pfreundschuh, and R.~Siegwart.
\newblock {Dynablox: Real-time Detection of Diverse Dynamic Objects in Complex
  Environments}.
\newblock {\em IEEE Robotics and Automation Letters (RA-L)}, 8(10):6259--6266,
  2023.

\bibitem{song2024neurips-gvkf}
G.~Song, C.~Cheng, and H.~Wang.
\newblock {GVKF: Gaussian Voxel Kernel Functions for Highly Efficient Surface
  Reconstruction in Open Scenes}.
\newblock In {\em Proc.~of the Conf. on Neural Information Processing Systems
  (NeurIPS)}, 2024.

\bibitem{stachniss2005rss}
C.~Stachniss, G.~Grisetti, and W.~Burgard.
\newblock {Information Gain-based Exploration Using Rao-Blackwellized Particle
  Filters}.
\newblock In {\em Proc.~of Robotics: Science and Systems (RSS)}, 2005.

\bibitem{sucar2021iccv}
E.~Sucar, S.~Liu, J.~Ortiz, and A.J. Davison.
\newblock imap: Implicit mapping and positioning in real-time.
\newblock In {\em Proc.~of the IEEE/CVF Intl.~Conf.~on Computer Vision (ICCV)},
  2021.

\bibitem{tancik2022cvpr-blocknerf}
M.~Tancik, V.~Casser, X.~Yan, S.~Pradhan, B.~Mildenhall, P.P. Srinivasan, J.T.
  Barron, and H.~Kretzschmar.
\newblock {Block-NeRF: Scalable Large Scene Neural View Synthesis}.
\newblock In {\em Proc.~of the IEEE/CVF Conf.~on Computer Vision and Pattern
  Recognition (CVPR)}, 2022.

\bibitem{tancik2023siggraph-nerfstudio}
M.~Tancik, E.~Weber, E.~Ng, R.~Li, B.~Yi, J.~Kerr, T.~Wang, A.~Kristoffersen,
  J.~Austin, K.~Salahi, A.~Ahuja, D.~McAllister, and A.~Kanazawa.
\newblock {Nerfstudio: A Modular Framework for Neural Radiance Field
  Development}.
\newblock In {\em Proc.~of the Intl.~Conf.~on Computer Graphics and Interactive
  Techniques (SIGGRAPH)}, 2023.

\bibitem{tao2024icra-silvr}
Y.~Tao, Y.~Bhalgat, L.F.T. Fu, M.~Mattamala, N.~Chebrolu, and M.~Fallon.
\newblock {SiLVR: Scalable Lidar-Visual Reconstruction with Neural Radiance
  Fields for Robotic Inspection}.
\newblock In {\em Proc.~of the IEEE Intl.~Conf.~on Robotics \& Automation
  (ICRA)}, 2024.

\bibitem{tao2024arxiv-oxford}
Y.~Tao, M.A. Munoz-Banon, L.~Zhang, J.~Wang, L.F.T. Fu, and M.~Fallon.
\newblock {The Oxford Spires Dataset: Benchmarking Large-Scale LiDAR-Visual
  Localisation, Reconstruction and Radiance Field Methods}.
\newblock {\em arXiv preprint}, arXiv:2411.10546, 2024.

\bibitem{thrun2001ai}
S.~Thrun, D.~Fox, W.~Burgard, and F.~Dellaert.
\newblock {Robust Monte Carlo Localization for Mobile Robots}.
\newblock {\em Artificial Intelligence}, 128(1-2), 2001.

\bibitem{vizzo2021icra}
I.~Vizzo, X.~Chen, N.~Chebrolu, J.~Behley, and C.~Stachniss.
\newblock {Poisson Surface Reconstruction for LiDAR Odometry and Mapping}.
\newblock In {\em Proc.~of the IEEE Intl.~Conf.~on Robotics \& Automation
  (ICRA)}, 2021.

\bibitem{vizzo2022sensors}
I.~Vizzo, T.~Guadagnino, J.~Behley, and C.~Stachniss.
\newblock {VDBFusion: Flexible and Efficient TSDF Integration of Range Sensor
  Data}.
\newblock {\em Sensors}, 22(3):1296, 2022.

\bibitem{vizzo2023ral}
I.~Vizzo, T.~Guadagnino, B.~Mersch, L.~Wiesmann, J.~Behley, and C.~Stachniss.
\newblock {KISS-ICP: In Defense of Point-to-Point ICP -- Simple, Accurate, and
  Robust Registration If Done the Right Way}.
\newblock {\em IEEE Robotics and Automation Letters (RA-L)}, 8(2):1029--1036,
  2023.

\bibitem{wang2021iros-fflo}
H.~Wang, C.~Wang, C.~Chen, and L.~Xie.
\newblock {F-LOAM: Fast LiDAR Odometry and Mapping}.
\newblock In {\em Proc.~of the IEEE/RSJ Intl.~Conf.~on Intelligent Robots and
  Systems (IROS)}, 2021.

\bibitem{wang2021neurips}
P.~Wang, L.~Liu, Y.~Liu, C.~Theobalt, T.~Komura, and W.~Wang.
\newblock Neus: Learning neural implicit surfaces by volume rendering for
  multi-view reconstruction.
\newblock In {\em Proc.~of the Conf.~on Neural Information Processing Systems
  (NeurIPS)}, 2021.

\bibitem{wang2004tip}
Z.~Wang, A.C. Bovik, H.R. Sheikh, and E.P. Simoncelli.
\newblock Image quality assessment: from error visibility to structural
  similarity.
\newblock {\em IEEE Transactions on Image Processing}, 13(4):600--612, 2004.

\bibitem{wei2024ral-gsfusion}
J.~Wei and S.~Leutenegger.
\newblock {GSfusion: Online RGB-D Mapping Where Gaussian Splatting Meets TSDF
  Fusion}.
\newblock {\em IEEE Robotics and Automation Letters (RA-L)},
  9(12):11865--11872, 2024.

\bibitem{whelan2015rss}
T.~Whelan, S.~Leutenegger, R.S. Moreno, B.~Glocker, and A.~Davison.
\newblock {ElasticFusion: Dense SLAM Without A Pose Graph}.
\newblock In {\em Proc.~of Robotics: Science and Systems (RSS)}, 2015.

\bibitem{wiesmann2023ral-locndf}
L.~Wiesmann, T.~Guadagnino, I.~Vizzo, N.~Zimmerman, Y.~Pan, H.~Kuang,
  J.~Behley, and C.~Stachniss.
\newblock {LocNDF: Neural Distance Field Mapping for Robot Localization}.
\newblock {\em IEEE Robotics and Automation Letters (RA-L)}, 8(8):4999--5006,
  2023.

\bibitem{wiesmann2024ral}
L.~Wiesmann, T.~L{\"a}be, L.~Nunes, J.~Behley, and C.~Stachniss.
\newblock {Joint Intrinsic and Extrinsic Calibration of Perception Systems
  Utilizing a Calibration Environment}.
\newblock {\em IEEE Robotics and Automation Letters (RA-L)}, 9(10):9103--9110,
  2024.

\bibitem{wiesmann2024arxiv-ba}
L.~Wiesmann, E.~Marks, S.~Gupta, T.~Guadagnino, J.~Behley, and C.~Stachniss.
\newblock {Efficient LiDAR Bundle Adjustment for Multi-Scan Alignment Utilizing
  Continuous-Time Trajectories}.
\newblock {\em arXiv preprint}, arXiv:2412.11760, 2024.

\bibitem{wu2024ral-vdbgpdf}
L.~Wu, C.L. Gentil, and T.~Vidal-Calleja.
\newblock {VDB-GPDF: Online Gaussian Process Distance Field with VDB
  Structure}.
\newblock {\em IEEE Robotics and Automation Letters (RA-L)}, 10(1):374--381,
  2024.

\bibitem{xie2024arxiv-gslivm}
Y.~Xie, Z.~Huang, J.~Wu, and J.~Ma.
\newblock {GS-LIVM: Real-Time Photo-Realistic LiDAR-Inertial-Visual Mapping
  with Gaussian Splatting}.
\newblock {\em arXiv preprint}, arXiv:2410.17084, 2024.

\bibitem{xu2022cvpr-pointnerf}
Q.~Xu, Z.~Xu, J.~Philip, S.~Bi, Z.~Shu, K.~Sunkavalli, and U.~Neumann.
\newblock {Point-NeRF: Point-Based Neural Radiance Fields}.
\newblock In {\em Proc.~of the IEEE/CVF Conf.~on Computer Vision and Pattern
  Recognition (CVPR)}, 2022.

\bibitem{yan2024eccv-streetgs}
Y.~Yan, H.~Lin, C.~Zhou, W.~Wang, H.~Sun, K.~Zhan, X.~Lang, X.~Zhou, and
  S.~Peng.
\newblock {Street Gaussians for Modeling Dynamic Urban Scenes}.
\newblock In {\em Proc.~of the Europ.~Conf.~on Computer Vision (ECCV)}, 2024.

\bibitem{yang2023cvpr-unisim}
Z.~Yang, Y.~Chen, J.~Wang, S.~Manivasagam, W.C. Ma, A.J. Yang, and R.~Urtasun.
\newblock {UniSim: A Neural Closed-Loop Sensor Simulator}.
\newblock In {\em Proc.~of the IEEE/CVF Conf.~on Computer Vision and Pattern
  Recognition (CVPR)}, 2023.

\bibitem{wang2019tog-dss}
W.~Yifan, F.~Serena, S.~Wu, C.~{\"{O}}ztireli, and O.~Sorkine{-}Hornung.
\newblock {Differentiable Surface Splatting for Point-based Geometry
  Processing}.
\newblock {\em ACM Trans.~on Graphics (TOG)}, 38(6):1--14, 2019.

\bibitem{yu2024neurips-gsdf}
M.~Yu, T.~Lu, L.~Xu, L.~Jiang, Y.~Xiangli, and B.~Dai.
\newblock {GSDF: 3DGS meets SDF for Improved Rendering and Reconstruction}.
\newblock In {\em Proc.~of the Conf. on Neural Information Processing Systems
  (NeurIPS)}, 2024.

\bibitem{yu2024tog-gof}
Z.~Yu, T.~Sattler, and A.~Geiger.
\newblock {Gaussian Opacity Fields: Efficient Adaptive Surface Reconstruction
  in Unbounded Scenes}.
\newblock {\em ACM Trans.~on Graphics (TOG)}, 43(6):1--13, 2024.

\bibitem{zhang2024arxiv-radegs}
B.~Zhang, C.~Fang, R.~Shrestha, Y.~Liang, X.~Long, and P.~Tan.
\newblock {RaDe-GS: Rasterizing Depth in Gaussian Splatting}.
\newblock {\em arXiv preprint}, arXiv:2406.01467, 2024.

\bibitem{zhang2024eccv-glorie}
G.~Zhang, E.~Sandstr{\"o}m, Y.~Zhang, M.~Patel, L.~Van~Gool, and M.R. Oswald.
\newblock {GLORIE-SLAM: Globally optimized rgb-only implicit encoding point
  cloud slam}.
\newblock In {\em Proc.~of the Europ.~Conf.~on Computer Vision (ECCV)}, 2024.

\bibitem{zhang2014rss}
J.~Zhang and S.~Singh.
\newblock {LOAM: Lidar Odometry and Mapping in Real-time}.
\newblock In {\em Proc.~of Robotics: Science and Systems (RSS)}, 2014.

\bibitem{zhang2018cvpr-lpips}
R.~Zhang, P.~Isola, A.A. Efros, E.~Shechtman, and O.~Wang.
\newblock {The Unreasonable Effectiveness of Deep Features as a Perceptual
  Metric}.
\newblock In {\em Proc.~of the IEEE/CVF Conf.~on Computer Vision and Pattern
  Recognition (CVPR)}, 2018.

\bibitem{zhang2018iros-evo}
Z.~Zhang and D.~Scaramuzza.
\newblock {A Tutorial on Quantitative Trajectory Evaluation for
  Visual(-Inertial) Odometry}.
\newblock In {\em Proc.~of the IEEE/RSJ Intl.~Conf.~on Intelligent Robots and
  Systems (IROS)}, 2018.

\bibitem{zhao2024eccv-tclcgs}
C.~Zhao, S.~Sun, R.~Wang, Y.~Guo, J.J. Wan, Z.~Huang, X.~Huang, Y.V. Chen, and
  L.~Ren.
\newblock {TCLC-GS: Tightly coupled lidar-camera gaussian splatting for
  surrounding autonomous driving scenes}.
\newblock In {\em Proc.~of the Europ.~Conf.~on Computer Vision (ECCV)}, 2024.

\bibitem{zhao2022rss-occslam}
L.~Zhao, Y.~Wang, and S.~Huang.
\newblock {Occupancy-SLAM: Simultaneously Optimizing Robot Poses and Continuous
  Occupancy Map}.
\newblock In {\em Proc.~of Robotics: Science and Systems (RSS)}, 2022.

\bibitem{zheng2024ral-gaussiangrasper}
Y.~Zheng, X.~Chen, Y.~Zheng, S.~Gu, R.~Yang, B.~Jin, P.~Li, C.~Zhong, Z.~Wang,
  L.~Liu, et~al.
\newblock {GaussianGrasper: 3D Language Gaussian Splatting for Open-vocabulary
  Robotic Grasping}.
\newblock {\em IEEE Robotics and Automation Letters (RA-L)}, 9(9):7827--7834,
  2024.

\bibitem{zhong2023icra}
X.~Zhong, Y.~Pan, J.~Behley, and C.~Stachniss.
\newblock {SHINE-Mapping: Large-Scale 3D Mapping Using Sparse Hierarchical
  Implicit Neural Representations}.
\newblock In {\em Proc.~of the IEEE Intl.~Conf.~on Robotics \& Automation
  (ICRA)}, 2023.

\bibitem{zhong2024cvpr}
X.~Zhong, Y.~Pan, C.~Stachniss, and J.~Behley.
\newblock {3D LiDAR Mapping in Dynamic Environments using a 4D Implicit Neural
  Representation}.
\newblock In {\em Proc.~of the IEEE/CVF Conf.~on Computer Vision and Pattern
  Recognition (CVPR)}, 2024.

\bibitem{zhou2024cvpr-drivinggs}
X.~Zhou, Z.~Lin, X.~Shan, Y.~Wang, D.~Sun, and M.H. Yang.
\newblock {DrivingGaussian: Composite Gaussian Splatting for Surrounding
  Dynamic Autonomous Driving Scenes}.
\newblock In {\em Proc.~of the IEEE/CVF Conf.~on Computer Vision and Pattern
  Recognition (CVPR)}, 2024.

\bibitem{zhu2025threedv-loopsplat}
L.~Zhu, Y.~Li, E.~Sandström, S.~Huang, K.~Schindler, and I.~Armeni.
\newblock {LoopSplat: Loop Closure by Registering 3D Gaussian Splats}.
\newblock In {\em Proc.~of the Intl.~Conf.~on 3D Vision (3DV)}, 2025.

\bibitem{zuo2024fmgs}
X.~Zuo, P.~Samangouei, Y.~Zhou, Y.~Di, and M.~Li.
\newblock {FMGS: Foundation model embedded 3d gaussian splatting for holistic
  3d scene understanding}.
\newblock {\em Intl.~Journal~of Computer Vision (IJCV)}, 130:1--17, 2024.

\bibitem{zwicker2001pv}
M.~Zwicker, H.~Pfister, J.~Van~Baar, and M.~Gross.
\newblock {EWA volume splatting}.
\newblock In {\em Proc.~of Visualization}, 2001.

\end{thebibliography}
